\documentclass[11pt]{article}

\usepackage[preprint]{acl}

\usepackage{times}
\usepackage{latexsym}
\usepackage{amsmath}

\usepackage[T1]{fontenc}
\usepackage[utf8]{inputenc}

\usepackage{microtype}

\usepackage{inconsolata}

\usepackage{graphicx}
\usepackage{stfloats}

\usepackage{iftex}
\ifPDFTeX
  \usepackage{CJKutf8}
\else
  \usepackage{xeCJK}
  \xeCJKDeclareCharClass{HalfRight}{"2019}
  \setCJKsansfont{FandolHei-Regular.otf}[BoldFont=FandolHei-Bold.otf]
  \setCJKmonofont{FandolFang-Regular.otf}
\fi

\usepackage{array}
\usepackage{longtable}
\usepackage[most]{tcolorbox}
\definecolor{promptgray}{RGB}{218,218,218}
\newtcolorbox{promptbox}[1]{
  enhanced,
  unbreakable,
  colback=black!1,
  colframe=promptgray,
  colbacktitle=promptgray,
  coltitle=black,
  fonttitle=\normalfont\bfseries\normalsize,
  fontupper=\normalfont\normalsize,
  title={#1},
  arc=1.2mm,
  boxrule=0.8pt,
  left=2.5mm,
  right=2.5mm,
  top=2mm,
  bottom=2mm,
  before skip=8pt,
  after skip=10pt,
  before upper={\setlength{\parindent}{0pt}\setlength{\parskip}{0.55\baselineskip}\raggedright}
}

\title{Are These Programmers Ripe Enough to Eat?\\Evaluating How LLMs Handle Lexical Knowledge Conflicts in Translation}

\author{Dongxu Wu \\
  The Ohio State University \\
  \texttt{wu.6734@osu.edu} \\\And
  Sara Court \\
  The Ohio State University \\
  \texttt{court.22@osu.edu} \\\And
  Micha Elsner \\
  The Ohio State University \\
  \texttt{elsner.14@osu.edu} \\}

\begin{document}
\ifPDFTeX\begin{CJK*}{UTF8}{gbsn}\fi
\maketitle
\begin{abstract}
External lexical resources can help large language models (LLMs) translate texts in low-resource languages and specialized domains, but these resources may conflict with models' prior lexical knowledge. To isolate this conflict, we construct a controlled Mandarin-to-English translation task in which familiar source words are assigned new meanings within and across semantic categories. We treat these prompt-supplied meanings as contextual knowledge (CK) and systematically evaluate the effects of word frequency, sentence context, and prompt framing on the translated outputs. Across nine models, CK reliance is greater for less frequent source words and in sentence contexts that support the supplied meanings. Prompt framing can reduce CK reliance, but seven of nine models favor CK even when explicitly instructed to ignore the supplied meanings. Within model families, larger models are more likely than smaller models to utilize CK in their outputs. Three follow-up experiments further show that modal weakening lowers CK reliance, the effect of repetition depends on prompt layout, and the effect of reasoning varies across models. Overall, our controlled study reveals how CK reliance varies across conditions and models, motivating new methods to detect and evaluate lexical conflicts and guide models' lexical choices in LLM-based translation.\footnote{All data and code are available at \url{https://anonymous.4open.science/r/lexical-conflict}.}

\end{abstract}

\begin{figure}[t]
  \centering
  \includegraphics[width=\columnwidth]{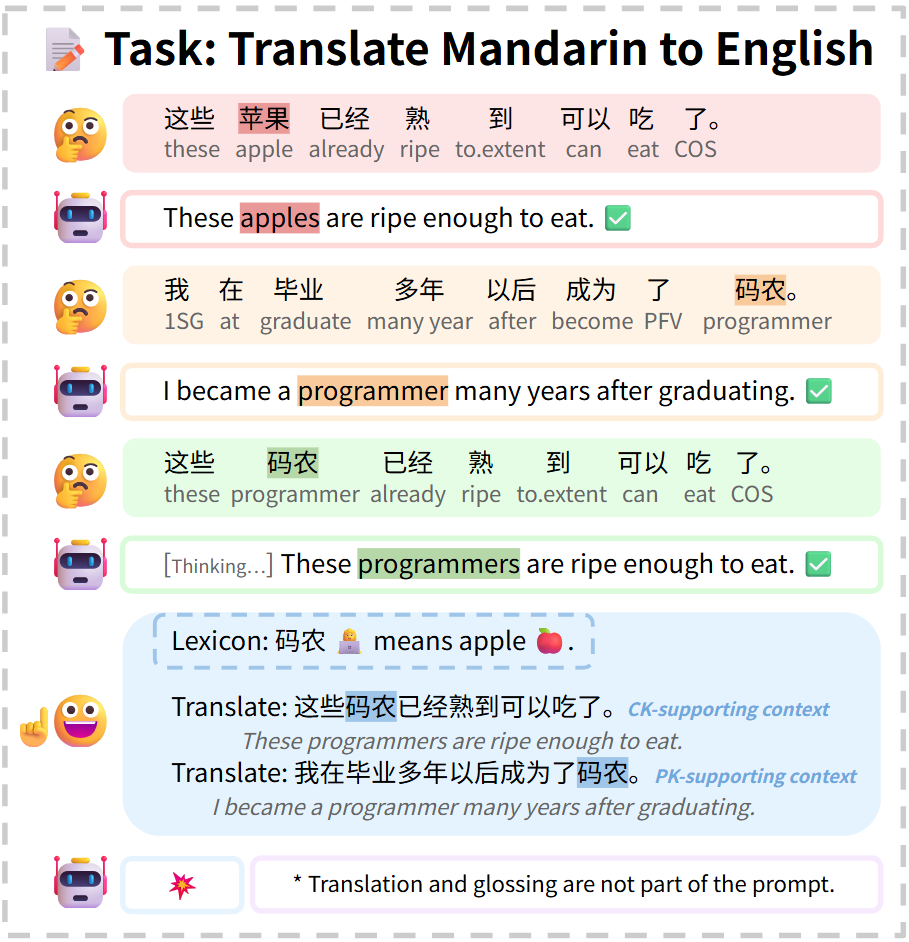}
  \caption{Schematic illustration of lexical conflicts in Mandarin-to-English translation. The prompt assigns the new meaning `apple' to the Mandarin word \textit{码农} \textit{mǎnóng} `programmer'. The model is then asked to translate sentences whose contexts favor either the familiar meaning or the new meaning.}
  \label{fig:task-example}
\end{figure}

\section{Introduction}
\label{sec:introduction}

LLM-based machine translation increasingly uses dictionaries and other external lexical resources to supplement models’ lexical knowledge, particularly for low-resource languages \citep{ghazvininejad2023dictionary,lu2025slow} and specialized domains \citep{bogoychev2023terminology,li2025domain}. However, some of the words covered by these resources may already be familiar to the model, and a supplied definition may conflict with the meaning the model learned during pretraining. This conflict may arise because the supplied definition is incorrect, because the model learned an incorrect or incomplete meaning, or because both meanings are valid, with the appropriate choice depending on the language variety, domain, or task context. We refer to this mismatch as \emph{lexical conflict} between \emph{parametric knowledge} (PK), the lexical knowledge encoded in the model’s parameters, and \emph{contextual knowledge} (CK), the word meanings supplied in the prompt.

Previous studies have documented lexical conflict as a contributor to errors in real translation settings \citep{huang2024translate,court2024shortcomings,merx2025tulun}. However, multiple factors vary together in real translation data, making their individual effects difficult to separate, and resource scarcity further limits experimental manipulation in low-resource languages \citep{barua2024lexical}.

We therefore take inspiration from work that uses carefully controlled synthetic data to study knowledge conflicts in question answering \citep{longpre2021entity} and sentence completion \citep{shi2024trusting}. Using Mandarin-to-English translation as a high-resource test case, our stimuli assign new English meanings to familiar Mandarin nouns drawn from five everyday semantic categories. For example, we redefine the word \emph{码农} \textit{mǎnóng} `programmer' to mean `apple' (Figure~\ref{fig:task-example}). The language pair and category set are deliberately chosen to reduce the risk that limited language proficiency or a lack of specialized domain knowledge will cause translation errors even without the new definitions, obscuring the effects of our manipulations. We examine three factors that often vary together in actual translation settings. First, we use corpus word frequency as a proxy for models' familiarity with Mandarin words and their conventional English translations, predicting lower CK reliance for more frequent words because of their stronger associations with familiar meanings. Second, we vary the sentence context to support either the familiar or the supplied meaning. For example, a sentence about ripeness supports `apple', whereas one about employment supports `programmer'. Third, we vary prompt framing: the same definition may appear as a bare dictionary entry, a rule for a newly documented variety, a language-game rule, or information that the model is explicitly instructed to accept or ignore.

We evaluate nine models on 150 two-character Mandarin nouns spanning five semantic categories and three frequency bands. Experiment~1 assigns new meanings across categories and varies sentence context across five prompt frames. Experiment~2 narrows the contrast to meanings within the same category. This resembles cases where a word has a different meaning in a regional variety or specialized domain, even though its usual meaning also fits the sentence. 
% As baselines for both experiments, we first ask each model to translate the same sentences without any lexical rule to check whether it produces the source word’s usual translation. We also test inclusion of the lexical rule using the target word's conventional meaning. 
Three follow-up experiments replace \emph{means} with \emph{may mean}, repeat the rule up to five times at different positions in the prompt, and enable reasoning in three models. The full lexical inventory and prompts appear in Appendices~\ref{app:lexical-materials} and~\ref{app:prompts}.

Across the nine models, CK reliance is greater for less frequent source words and in sentence contexts that support the supplied meanings. It remains high even when both meanings are compatible with the sentence context, indicating that CK adoption cannot be explained solely by the repair of implausible sentences. Within
% the paired open-weight 
model families, larger models also rely more on CK.
% A proxy for target-word frequency shows no comparably robust or consistent effect. 
Prompt framing shapes lexical choices: in both main experiments, a bare dictionary entry elicits more CK responses than the same rule framed as a definition for a newly documented variety or a language game. Explicit instructions shift these choices, yet seven of nine models favor CK even when instructed to ignore the supplied meanings. The follow-up experiments show that replacing \emph{means} with \emph{may mean} lowers CK reliance by 17.73 percentage points overall, the effect of repeating the rule \textit{n} times depends on where the \textit{n} copies are placed in the prompt, and the effect of reasoning varies across models. Enabling reasoning improves GPT-5.5’s compliance with the ignore instruction while driving both Qwen models toward near-universal CK adoption.

\section{Related Work}
\label{sec:related-work}

\subsection{Knowledge Conflicts in LLMs}
\label{sec:related-pk-ck}

Previous studies on knowledge conflicts for question answering test how models respond when supplied with information that contradicts a familiar answer. \citet{longpre2021entity} replaced answer entities in question-answering passages and found that generative readers sometimes retained previously memorized answers. \citet{xie2024adaptive} found greater adoption of a conflicting answer with generated contradictory passages than with entity-substituted passages, showing that the construction of the evidence also matters. DisentQA uses counterfactual and unanswerable examples to train models to give separate answers based on the supplied context and their parametric knowledge \citep{neeman2023disentqa}.

Familiar-word redefinition has also been studied outside translation. The Redefine task tests whether language models can follow redefinitions of familiar words and symbols when answering simple questions \citep{mckenzie2023inverse}. WinoDict includes a meaning-shift condition that assigns new definitions to familiar words; this was harder for PaLM-540B than its synthetic-word condition \citep{eisenschlos2023winodict}. In matched English lexical-remapping prompts, \citet{wang2026persistent} found that familiar associations continued to affect the relative probabilities of target and distractor responses after a local redefinition. Our translation task systematically examines how this competition affects models' translations of the target words when prompted to translate full sentences from Mandarin to English.

\subsection{External Lexical Knowledge in LLM Translation}
\label{sec:related-dictionary}

Dictionary prompting supplies candidate translations that a model can use in its output. For example, \citet{ghazvininejad2023dictionary} improved low-resource and domain-specific translation with bilingual hints. 
% Later studies have shown that
% the format of these hints 
Format also matters: \citet{lu2024chain} linked equivalent words across several languages in dictionary chains, outperforming the same multilingual information presented as separate word pairs.
% in their ablation.

Linguistic documentation can provide a broader form of support for extremely 
low-resource translation. 
\citet{tanzer2024grammar} introduced MTOB, a Kalamang--English benchmark using a grammar book, word list, and parallel examples. However, 
% by separating grammar explanations from parallel material, 
\citet{aycock2025grammar} found that the parallel material accounted for most translation gains in Kalamang, with similar results for Nepali and Guarani.

A major issue facing such methods is that automated retrieval 
% from external resources
% is error-prone, and 
can 
% unintentionally 
provide incorrect or misleading information to the downstream model. Several previous papers report lexical conflict as a source of errors. 
% In Southern Quechua-to-Spanish translation, \citet{court2024shortcomings} found that morpheme information helped some models but reduced GPT-4o's BLEURT score relative to its baseline. 
For example, \citet{merx2025tulun} documented contextually inappropriate glossary application during post-editing. \citet{huang2024translate} report cases in which LLM translations favor common lexical renderings over supplied terminology constraints. \citet{dinatale2025legistyr} also evaluated legal translation from Italian into South Tyrolean German, finding that their best model used supplied target terms only 64.5\% of the time; they also found errors in contextual disambiguation.

\subsection{Factors Affecting Lexical Choices}
\label{sec:related-factors}

External resources can support a model's ability to process unfamiliar words, but may harm performance when they conflict with its knowledge of more familiar words.
% Corpus word frequency provides a practical proxy for lexical familiarity: 
% In human word processing, frequency reflects exposure, although estimates depend on the corpus \citep{brysbaert2018frequency}. 
\citet{lu2025slow} used public frequency estimates to prioritize entries for less frequent words, improving translation over alternative selection methods under a fixed entry budget. 
% Knowledge-conflict studies suggest why familiarity may matter. For example, in a country--capital task, \citet{yu2023factual} found that Pythia models more often retained a memorized capital when the country occurred more frequently in their training corpus. 
% \citet{du2024context} found less context-induced change for entity--answer relations with more training co-occurrences. 
We test a related prediction for our translation task: models should be less likely to adopt supplied meanings for more frequent source words.

When a word has multiple senses, sentence context can favor one lexical interpretation over another. \citet{barua2024lexical} introduced DTAiLS, in which source sentences provide contextual cues for choosing among target-language alternatives. They also found that natural-language rules describing these distinctions
% , generated by GPT-4 and checked by native speakers, 
improved lexical selection. In our redefinition task, we vary whether the sentence supports the familiar or supplied meaning to test how semantic support affects CK reliance.

How an assertion is presented can also affect whether a model follows it. \citet{zhou2023contextfaithful} increased use of contextual answers by attributing a passage to a narrator and asking for that narrator's view. In multiple-choice question answering, models relied more on assertions attributed to documents than on those attributed to users in most tested model--dataset combinations \citep{li2026balance}. We examine this sensitivity to presentation by varying the framing around the same lexical definition, including instructions to use or ignore it. 
% The certainty of the assertion provides a related basis for variation: \citet{du2026belief} found relatively low context following when factual claims were expressed tentatively. Our follow-up experiment tests this issue by replacing \emph{means} with \emph{may mean}.

\begin{table}[t]
  \centering
  \normalsize
  \setlength{\tabcolsep}{6pt}
  \begin{tabular}{@{}lp{6.45cm}@{}}
    \hline
    \textbf{ID} & \textbf{Frame template} \\
    \hline
    F1 & Use what you know about Mandarin and ignore any newly created lexical rule. [SOURCE] means [TARGET]. \\
    F2 & Lexicon: [SOURCE] means [TARGET]. \\
    F3 & We are using a newly documented variety of Mandarin. In this variety, [SOURCE] means [TARGET]. \\
    F4 & We are playing a language game. In this game, [SOURCE] means [TARGET]. \\
    F5 & Use the lexicon below even if it conflicts with standard Mandarin. [SOURCE] means [TARGET]. \\
    \hline
  \end{tabular}
  \caption{Prompt-frame templates used in the main experiments. [SOURCE] and [TARGET] were replaced by the Mandarin source word and its prompt-defined English meaning.}
  \label{tab:pragmatic-frames}
\end{table}

\section{Experimental Setup}
\label{sec:experimental-setup}

\subsection{Materials and Conditions}
\label{sec:materials-conditions}

We use a Mandarin-to-English translation task with 150 two-character Mandarin nouns from five everyday semantic categories: animals, fruits, vehicles, occupations, and musical instruments. 
% This language pair and category set are chosen specifically to reduce the risk that limitations in models' language proficiency may cause them to fail even on the baseline task and further obscure the effects we hope to see from our manipulations.
Within each category, we rank candidate nouns by their BCC corpus frequency \citep{xun2016bcc} and select ten words from each of three bands: high, medium, and low. Appendix~\ref{app:lexical-materials} gives the selection procedure and inventory. Source frequency refers to the Mandarin word being redefined. For the supplied English meaning, we use the BCC frequency band of its Mandarin equivalent as a target-frequency proxy; these bands may differ from English frequency rankings, but provide insight into the model's familiarity with the conventional Mandarin meaning of the target word's new English definition.

We use one sentence template per category. In Experiment~1, each mapping assigns a meaning from another category and appears in two contexts: one supports the familiar meaning (PK context), and the other supports the supplied meaning (CK context). In Experiment~2, both meanings belong to the same category and fit its sentence template. Both experiments use five prompt frames, varying how the definition is introduced and whether the model is instructed to use or ignore it (Table~\ref{tab:pragmatic-frames}). Appendix~\ref{app:prompts} gives the exact prompts.

\subsection{Models and Response Coding}
\label{sec:models-coding}

We evaluate GPT-5.5 \citep{openai2026gpt55} and four open-weight model families, each represented by a smaller and a larger model: Qwen3-8B and Qwen3-32B \citep{yang2025qwen3}; Llama-3-8B and Llama-3-70B \citep{llama3modelcard}; DeepSeek-LLM-7B and DeepSeek-LLM-67B \citep{deepseekai2024deepseekllm}; and Gemma-3-4B and Gemma-3-27B \citep{gemmateam2025gemma3}. GPT-5.5's reasoning mode and Qwen3's thinking mode are disabled in the main experiments. We retain one complete, non-empty response per prompt. Appendix~\ref{app:inference-details} documents generation settings and recovery requests.

We code only the final response text, excluding any separately returned thinking or reasoning content. We automatically match each response against accepted English forms for the familiar and supplied meanings, including inflections and conventional synonyms (Appendix~\ref{app:lexical-materials}). Responses are coded as \textsc{PK} when only a familiar form appears, \textsc{CK} when only a supplied form appears, \textsc{Both} when both appear, and \textsc{Other} when neither appears. We measure CK reliance as the proportion of valid responses coded as \textsc{CK}; \textsc{PK}, \textsc{Both}, and \textsc{Other} remain in the denominator.

\begin{figure}[t]
  \centering
  \includegraphics[width=\columnwidth]{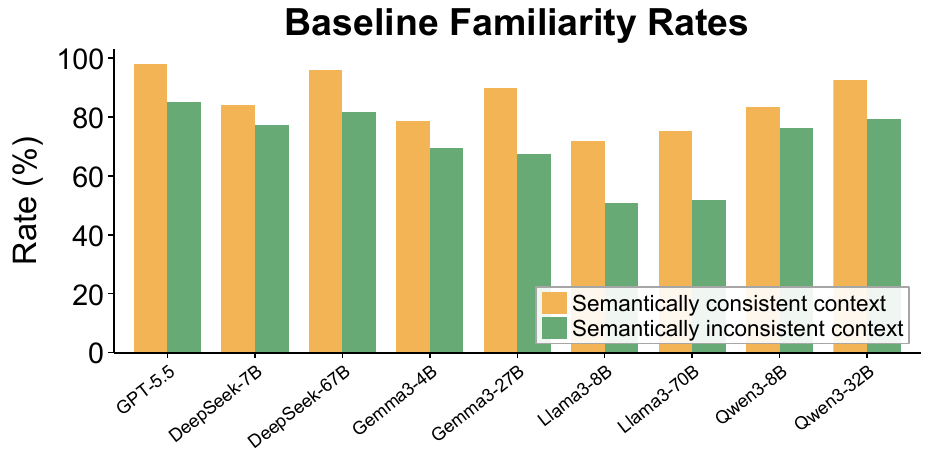}
  \caption{Baseline familiarity rates by model. For each model, the left and right bars show the proportion of \textsc{Familiar} responses in semantically consistent and inconsistent contexts, respectively.}
  \label{fig:baseline-rates}
\end{figure}

\subsection{Baselines and Analysis}
\label{sec:baselines-filtering}
\label{sec:statistical-analysis}

An implausible sentence can itself affect lexical choice, so our no-rule baselines test whether each model translates the source word's familiar meaning without the new definition. We use the template for the source word's own category (150 semantically consistent prompts per model) and the other four templates (600 semantically inconsistent prompts) (Appendix~\ref{app:baseline-prompts}). A response is \textsc{Familiar} if it contains an accepted translation of the source word's familiar, conventional meaning. The familiarity rate fell from 85.56\% in semantically consistent contexts to 71.00\% in semantically inconsistent contexts, with a decrease in every model (Figure~\ref{fig:baseline-rates}). These baselines show that models do not always produce the familiar translation and that sentence context alone can affect lexical choice. An additional baseline shows that a lexical rule agreeing with the familiar meaning raises familiar-form inclusion to near ceiling (Appendix~\ref{app:paper-conventional-meaning}).

Primary analyses use all valid responses. Robustness analyses retain only trials whose matched baseline is \textsc{Familiar}, with matches defined by model, source word, and sentence template (Appendices~\ref{app:filtered-robustness} and~\ref{app:followup-results}).
% This checks whether the main findings hold for cases where the model correctly translates the source word's familiar meaning without any rule.

We fit separate binomial logistic mixed-effects models for Experiments~1 and~2, predicting \textsc{CK} (1) versus all other responses (0). The models use additive fixed effects for frame, source frequency, and the target-frequency proxy, with context included only in Experiment~1. All predictors are categorical. Random intercepts for LLM identity, source category, source item, target item, and mapping account for repeated observations. We report odds ratios, unadjusted 95\% Wald intervals, and Holm-adjusted $p$ values \citep{holm1979sequential}; Appendix~\ref{app:analysis-audits} gives formulas and correction families. Follow-ups, model-size comparisons, and cross-experiment comparisons are descriptive.

\begin{figure*}[!t]
  \centering
  \includegraphics[width=\textwidth]{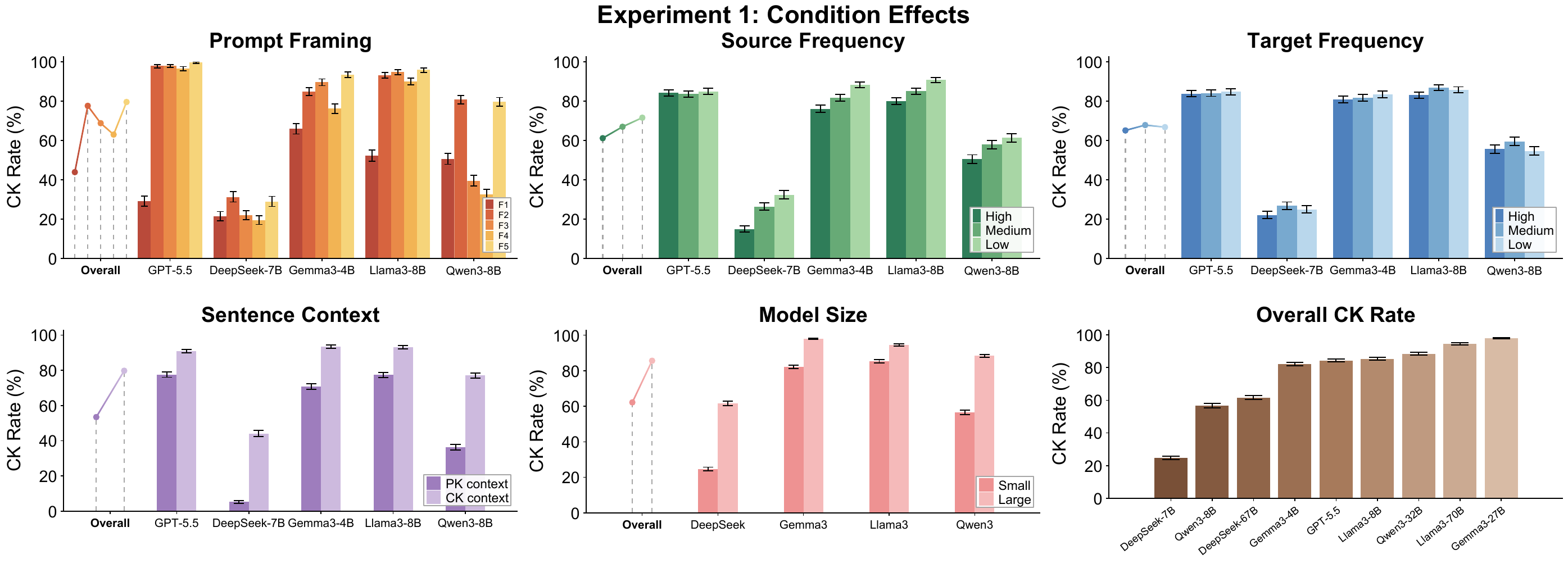}
  \caption{Experiment~1 CK rates in the full dataset, by prompt frame, source frequency, target-frequency proxy, sentence context, model size, and model. The first four panels show GPT-5.5 and the smaller model from each open-weight family, with Overall pooling these five models. Full results for all nine models appear in Appendix~\ref{app:exp1-full-results}.}
  \label{fig:exp1-results}
\end{figure*}

\section{Experiment 1: Lexical Conflicts across Semantic Categories}
\label{sec:exp1}

\subsection{Motivation and Design}
\label{sec:exp1-design}

Experiment~1 tests how word frequency, sentence context, and prompt framing affect the use of a conflicting lexical definition. Each mapping assigns a source noun a meaning from another semantic category. Every mapping appears under all five prompt frames and in two sentences: one supports the familiar meaning (PK context), and the other supports the supplied meaning (CK context).

\subsection{Results}
\label{sec:exp1-results}

\paragraph{Overall lexical choices.}
Across models and conditions (Figure~\ref{fig:exp1-results}), 75.07\% of responses were \textsc{CK}, compared with 19.01\% \textsc{PK}, 0.60\% \textsc{Both}, and 5.32\% \textsc{Other}. Models therefore used the supplied meaning in most responses, even when it belonged to a different semantic category.

\paragraph{Sentence context.}
CK rates increased from 64.29\% in PK contexts to 85.86\% in CK contexts. In the mixed model, CK contexts increased the odds of a CK response by a factor of 8.26 (95\% CI [7.77, 8.79]; $\chi^2(1)=4520.57$, $p<.001$). Sentence context substantially shifted lexical choices, although CK remained common when the sentence supported the familiar meaning.

\paragraph{Word frequency.}
For source words, CK rates rose as frequency declined (high: 70.92\%; medium: 75.33\%; low: 78.97\%). The frequency effect was significant, $\chi^2(2)=55.23$, $p<.001$. Relative to high-frequency words, CK odds were higher for medium-frequency (OR $=1.48$, 95\% CI [1.21, 1.81]) and low-frequency words (OR $=2.12$, 95\% CI [1.74, 2.59]); both contrasts had $p<.001$. Rates across the target-frequency proxy bands ranged from 73.96\% to 75.99\%, with no significant effect, $\chi^2(2)=1.79$, $p=.408$.

\paragraph{Prompt framing.}
Frame strongly affected CK responses, $\chi^2(4)=3337.14$, $p<.001$. CK accounted for a majority of responses under F1's instruction to ignore the new rule (59.70\%) and was highest under F5's instruction to follow it (84.46\%). Among the other frames, CK rates were highest in F2, followed by F3 and F4 (81.96\%, 77.52\%, and 71.72\%). A bare dictionary entry thus elicited more CK responses than the same definition presented as a rule for a newly documented variety or a language game.

\paragraph{Model variation.}
CK rates ranged from 24.65\% for DeepSeek-LLM-7B to 97.97\% for Gemma-3-27B. Within every paired open-weight family, the larger model had a higher CK rate. Larger models usually showed smaller differences across conditions, although DeepSeek-LLM-67B had larger context and frame differences than its 7B counterpart. Near-ceiling CK rates were also found for the larger models.
% also compress differences on the probability scale.

\section{Experiment 2: Lexical Conflicts within Semantic Categories}
\label{sec:exp2}

\subsection{Motivation and Design}
\label{sec:exp2-design}

In Experiment~1's CK contexts, adopting the supplied meaning can make an otherwise implausible sentence plausible. Experiment~2 tests whether models still adopt that meaning when both meanings fit the sentence. This resembles cases where a word has a different meaning in a regional variety or specialized domain, while its usual meaning also fits the sentence. Each mapping pairs words from different frequency bands within the same semantic category and appears in one sentence context. We retain Experiment~1's five prompt frames.

\subsection{Results}
\label{sec:exp2-results}

\paragraph{Overall lexical choices.}
CK accounted for 79.58\% of responses (Figure~\ref{fig:exp2-results}); 17.34\% were \textsc{PK}, 0.31\% \textsc{Both}, and 2.77\% \textsc{Other}. Models therefore continued to adopt supplied meanings when the familiar meaning also fit the sentence.

\paragraph{Word frequency.}
CK rates rose as source frequency decreased (high: 76.13\%; medium: 79.47\%; low: 83.14\%; $\chi^2(2)=25.59$, $p<.001$). Relative to high-frequency words, ORs were 1.79 for medium-frequency words (95\% CI [1.27, 2.54], $p=.003$) and 2.42 for low-frequency words (95\% CI [1.71, 3.43], $p<.001$). The target-frequency proxy again had no significant effect ($\chi^2(2)=3.50$, $p=.174$).

\paragraph{Prompt framing.}
The frame effect remained strong, $\chi^2(4)=1921.78$, $p<.001$. CK rates ranged from 63.39\% in F1 to 87.87\% in F5. As in Experiment~1, rates among the other frames were highest in F2, followed by F3 and F4 (87.43\%, 81.96\%, and 77.26\%). In both experiments, seven of nine models produced more than 50\% CK responses despite the instruction to ignore the rule.

\paragraph{Model variation and comparison with Experiment 1.}
CK rates ranged from 27.93\% for DeepSeek-LLM-7B to 99.73\% for Gemma-3-27B. The larger model again had a higher CK rate in each paired open-weight family. The pooled rate was 4.51 percentage points higher than in Experiment~1, with increases in all five frames and eight of nine models. Appendix~\ref{app:exp2-full-results} gives the descriptive comparison across experiments.

\paragraph{Baseline-filtered results.}
After retaining only trials with a \textsc{Familiar} response in the matched no-rule baseline, CK rates were 71.85\% in Experiment~1 and 77.80\% in Experiment~2. Framing and source frequency remained significant in both experiments, and sentence context remained significant in Experiment~1. The target-frequency proxy remained nonsignificant in both (Appendix~\ref{app:filtered-robustness}). CK adoption therefore remained common among cases where the model produced the familiar translation without the new rule.

\begin{figure}[htbp]
  \centering
  \includegraphics[width=\columnwidth]{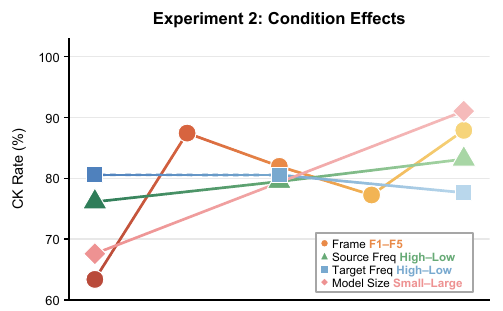}
  \caption{Experiment~2 CK rates in the full dataset, pooled across models for prompt framing, source frequency, the target-frequency proxy, and model size.}
  \label{fig:exp2-results}
\end{figure}

\begin{figure*}[!t]
  \centering
  \includegraphics[width=\textwidth]{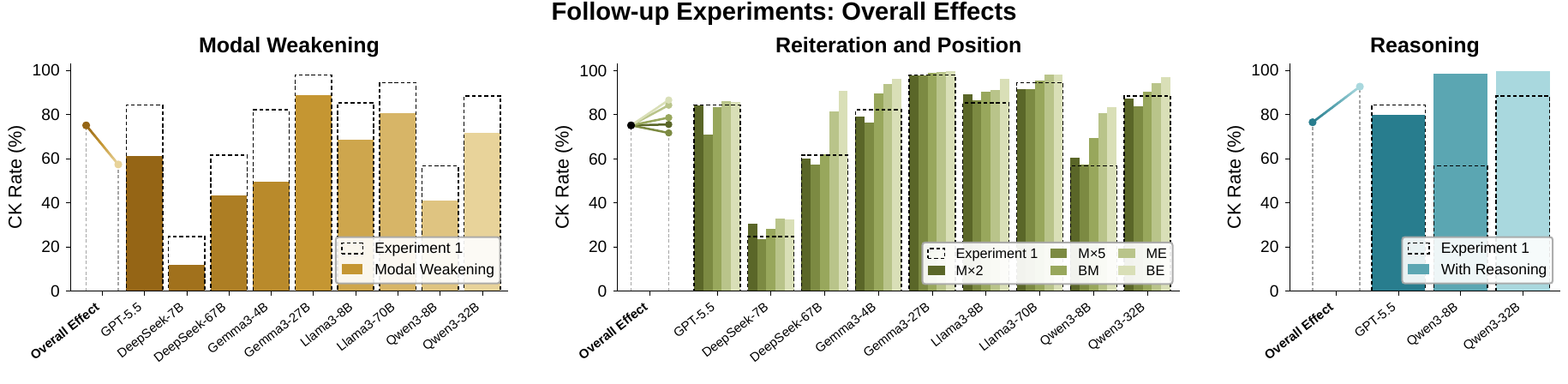}
  \caption{Model-level and pooled CK rates in the three follow-ups, compared with Experiment~1 (dashed outlines). The Reiteration and Position panel shows five layouts per model; $\times n$ denotes $n$ successive copies at M (Appendix~\ref{app:paper-position-repetition}). Slopes on the left show changes pooled across the models in each panel.}
  \label{fig:followups-overall}

  \par\vspace{\dblfloatsep}
  \centering
  \includegraphics[width=\textwidth]{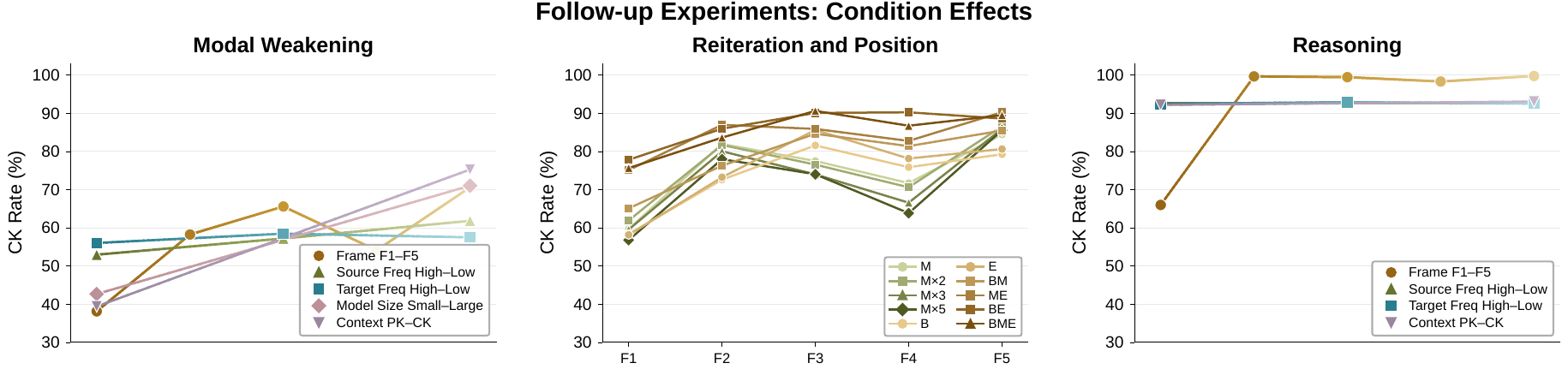}
  \caption{Condition-specific CK rates in the three follow-ups, pooled across models. The Reiteration and Position panel shows all ten layouts by prompt frame. Middle-only layouts (green) have one, two, three, or five copies; layouts involving B or E are brown.}
  \label{fig:followups-conditions}
\end{figure*}

\section{Controlled Follow-up Experiments}
\label{sec:followups}

The main experiments show that models often use a supplied definition, even when instructed to ignore it. The follow-ups test whether this reliance changes when the definition is tentative, when it is repeated throughout the prompt, or when reasoning is enabled. Each uses the mappings, frames, sentence contexts, and response coding of Experiment~1. Figures~\ref{fig:followups-overall} and~\ref{fig:followups-conditions} show overall and condition-specific CK rates for the three follow-ups.

\subsection{Modal Weakening}
\label{sec:modal-weakening}

\paragraph{Motivation and design.}
A retrieved entry may list several possible meanings, some correct and others incorrect or unsuitable for the sentence. Tentative wording can signal this uncertainty. We therefore test whether models respond to it by replacing \emph{means} with \emph{may mean} in every Experiment~1 prompt. If models treat the latter as a weaker claim, they should be less likely to use the supplied meaning.

\paragraph{Results.}
Modal weakening reduced the pooled CK rate from 75.07\% to 57.34\%, with declines of 9.33--32.70 percentage points across all models.

The context effect increased substantially: the CK-rate gap between CK and PK contexts widened from 21.56 percentage points in Experiment~1 (85.86\% vs.\ 64.29\%) to 35.87 points under modal weakening (75.28\% vs.\ 39.41\%); CK declined more sharply in PK contexts than in CK contexts (24.88 vs.\ 10.58 percentage points). This suggests that modal weakening makes models' adoption of the supplied meaning more dependent on whether it is supported by the sentence context.

The remaining condition patterns persisted: CK increased as source frequency declined (52.97\%, 57.22\%, and 61.84\% for high, medium, and low frequency). Across frames, CK ranged from 38.14\% in F1 to 70.61\% in F5.

\subsection{Lexical-Rule Reiteration and Position}
\label{sec:reiteration}

\paragraph{Motivation and design.}
Repeating a rule is a simple way to emphasize it. For example, \citet{sun2026taskmatters} found that repeating contradictory passages could increase model reliance on their contents, with the consequences depending on which knowledge source the task required. \citet{schuster2026source} found that repeating one of two conflicting tables could reverse which source the model favored. Studies on long-context processing also 
indicate that the position of relevant information matters: models can have difficulty using information located in the middle of long contexts \citep{li2025long,liuetal-2024-lost}. We therefore test whether repeating the lexical rule increases its use in the resulting translation, and compare prompt layouts in which the rule may be included before the task instruction (B), between the task instruction and the source sentence (M), and/or after the source sentence (E). In position M, the placement used in the main experiments, we test $n \in \{1, 2, 3, 5\}$ repetitions in direct succession for each Experiment~1 prompt. The frame-specific wording such as \emph{In this variety} in every copy is retained. Appendix~\ref{app:reiteration-prompts} specifies the layouts and Appendix~\ref{app:paper-position-repetition} gives complete results. 

\paragraph{Results.}
Repeating the rule in the middle position did not consistently increase CK: pooled rates were 75.07\%, 75.48\%, 73.23\%, and 71.65\% for one, two, three, and five copies (Figure~\ref{fig:followups-conditions}). With five copies, seven models decreased, while Llama-3-8B and Qwen3-8B increased by 1.07 and 0.78 points. In general, placement matters more than count. Adding a copy after the source raised CK to 84.22\% (ME), and placing the two copies before the task and after the source raised it to 86.56\% (BE), compared with 75.48\% for M$\times$2; BE exceeded M$\times$2 in all nine models (Figure~\ref{fig:followups-overall}) and in every frame, including F1, where CK rose from 61.94\% to 77.83\% despite the instruction to ignore the rule. Three distributed copies (BME, 85.24\%) did not add to the two-copy effect.

\subsection{Explicit Reasoning and Thinking Mode}
\label{sec:reasoning-followup}

\paragraph{Motivation and design.}
Reasoning models are increasingly common and may offer stronger instruction following. Additional reasoning may therefore help a model distinguish an instruction to ignore a rule from one to follow it. F1 provides a direct test: improved adherence would reduce CK responses under the ignore instruction. We reran the 6,000 Experiment~1 prompts with medium reasoning for GPT-5.5 and thinking mode for Qwen3-8B and Qwen3-32B. Each model is compared with its own Experiment~1 responses. Appendix~\ref{app:inference-details} documents the reasoning and output budgets.

\paragraph{Results.}
Enabling reasoning changed CK reliance in opposite directions for GPT-5.5 and the two Qwen models. GPT-5.5's overall CK rate decreased from 84.32\% to 79.77\%. Under F1, CK fell from 29.08\% to 0\% and PK rose from 67.67\% to 97.00\%, showing greater adherence to the ignore instruction. CK in the other four frames remained near ceiling (99.67--99.75\%). Qwen3-8B's CK rate increased from 56.68\% to 98.63\%, and Qwen3-32B's from 88.43\% to 99.55\%: both models used the supplied meaning on almost every trial.

\paragraph{Baseline-filtered results.}
Across all follow-ups, baseline filtering preserved the direction of each model's change in CK reliance (Appendices~\ref{app:followup-results} and~\ref{app:paper-position-repetition}).

\section{Discussion}
\label{sec:discussion}

The main experiments and follow-ups show that models respond to supplied definitions, but do not consistently use them as instructed. Without reasoning enabled, no prompt frame we tested guarantees that a model will use a supplied definition. This variability is to be assumed when using retrieved materials to support models for low-resource machine translation.

Source frequency and sentence context help explain the competition between familiar and supplied meanings. More frequent words elicited fewer CK responses in both main experiments, consistent with stronger associations with their familiar meanings. Sentence context also shaped lexical choices: models adopted supplied meanings more often when the sentence supported them. Yet CK adoption remained common in Experiment~2, where both meanings fit the sentence. Semantic support therefore influences the outcome, while adoption of a supplied meaning extends beyond repairing an implausible sentence.

Prompt framing adds a pragmatic dimension by changing the status of the definition and the instruction about its use. A bare dictionary entry elicited more CK responses than the same definition introduced as a rule for a newly documented variety or a language game. The follow-ups further distinguish the effects of certainty and repetition: replacing \emph{means} with \emph{may mean} reduced CK adoption across all nine models, consistent with \citet{du2026belief}, who found relatively low context following when conflicting factual claims were expressed tentatively. Repeating the definition up to five times in one place did not consistently increase it. Position, however, played an important role: two copies distributed before the task and after the source text elicited more CK than any other condition. Models thus respond to how a meaning is presented, but elaborating or repeating lexical guidance cannot be assumed to strengthen its effect.

Sensitivity to wording also needs to be distinguished from instruction adherence. A CK response follows F5's request to use the definition and violates F1's request to ignore it. Larger models' higher CK rates therefore do not by themselves establish better handling of lexical conflict. Reasoning further illustrates this distinction: GPT-5.5 followed the ignore instruction more consistently while continuing to adopt supplied meanings under the other frames, whereas both Qwen models adopted supplied meanings almost universally, including under the ignore instruction. The effects of prompt wording and reasoning settings therefore need to be evaluated against the required lexical choice for each model.

Our study uses a high-resource language pair and everyday semantic categories specifically to reduce the risk that limitations in language proficiency or specialized domain knowledge would cause baseline translation failures and obscure the effects of our manipulations. Even in this setting, CK reliance varied substantially across models and conditions, and supplied definitions were not consistently used as instructed. Our work thus complements existing research on retrieval-augmented generation for machine translation by providing a carefully controlled study of the factors that matter when using external resources to supplement model knowledge of a language. 

For practical translation, our findings favor omitting lexical entries known to be unsuitable for the intended translation. This recommendation is consistent with evidence that irrelevant context can impair reasoning \citep{shi2023distracted} and that an incorrect answer can bias output even when explicitly labeled as unrelated text \citep{kortukov2024documents}. Including an unwanted meaning and asking the model to ignore it may still lead the model to use that meaning. Evaluations should therefore check both whether requested meanings are used and whether unwanted meanings still appear in the output.

\section*{Limitations}

The experiments use synthetic word redefinitions in Mandarin-to-English translation covering only nouns. We intentionally select high-resource languages and everyday semantic categories to maximize the models' familiarity with our baseline stimuli for a stable no-rule comparison. This coverage limits how far the findings can be generalized to other language pairs, parts of speech, and conflicts arising from authentic lexical resources. Each category has only one sentence template, so the observed context effects may depend on the particular sentences chosen. Category assignments and sentence plausibility were not independently normed, leaving the strength and consistency of the intended semantic contrasts uncertain.

Word frequency comes from the BCC corpus and may differ from exposure in each model's training data. The observed frequency effect therefore provides indirect evidence about learned lexical associations. The target-frequency proxy uses counts for the Mandarin equivalents of the supplied English words, whose frequency ordering may differ across languages. Its nonsignificant effect does not establish that English target-word frequency has no influence.

The five prompt frames jointly vary instructions, discourse settings, and the stated scope of the definition, so their effects cannot be attributed separately to these components. Modal weakening uses only the contrast between \emph{means} and \emph{may mean}, leaving its sensitivity to other expressions of uncertainty untested. The reasoning comparison covers GPT-5.5 and two Qwen sizes with different reasoning implementations and output budgets, so differences between models cannot be attributed solely to how they reason.

The coding scheme tracks accepted English forms, so a \textsc{PK} or \textsc{CK} label does not establish that the full sentence was translated accurately. We provide COMET and chrF++ scores in Appendix~\ref{app:translation-quality} for reference; detailed analysis of the qualitative effects of lexical conflict on model outputs is beyond the scope of this paper. Paraphrases and outputs blending the familiar and supplied meanings, among other response types, are coded as \textsc{Other} when no accepted form for either meaning appears. This can group valid renderings outside the accepted lists with unsuccessful translations. We did not count these subtypes separately, so their contributions to the \textsc{Other} rate cannot be distinguished.

\section*{Generative AI Usage Statement}

OpenAI Codex and ChatGPT were used to assist with constructing experimental prompts and lexical mappings, generating accepted English form lists, implementing and auditing experimental and evaluation code, language polishing, and LaTeX editing. The authors have reviewed and verified all AI-assisted outputs and take full responsibility for the accuracy, originality, and integrity of the work.

% Camera-ready only:
% \section*{Acknowledgments}
% Acknowledgments are omitted from the anonymous review version.


\bibliography{custom}

\appendix
\section{Lexical Materials}
\label{app:lexical-materials}

The initial pool contained 60 two-character Mandarin nouns per semantic category. Items had to be attested in the BCC frequency list, fit the category template naturally, have a stable English translation, avoid strong polysemy, and have clear category membership. Frequencies use the 403,819,602-token BCC multi-domain corpus. For raw count $c$, frequency per million is $c/403{,}819{,}602 \times 10^6$, and the transformed measure is $\log_{10}(\text{frequency per million}+1)$. After ranking the 60 candidates within each category, ranks 1--10, 26--35, and 51--60 formed the high-, medium-, and low-frequency bands, respectively. The gaps between selected ranks make the bands more distinct.

Automatic coding matches the final response against accepted English forms, including the canonical gloss, singular/plural variants, and conventional synonyms. Matching is case-insensitive. Responses matching both the familiar and supplied form sets receive \textsc{Both}; those matching neither receive \textsc{Other}. Table~\ref{tab:complete-lexical-inventory} lists all accepted forms alongside each item's category, corpus frequency measure, and frequency band.

\onecolumn
% Appendix A lexical table. Prose, page breaks and column switches are in acl_latex.tex.
% Generated from data/processed/items.csv and data/coding/accepted_translations.csv.
% One continuous full-width multipage table; page breaks are handled by longtable.
{
\normalsize
\setlength{\tabcolsep}{2pt}
\renewcommand{\arraystretch}{1.0}
\setlength{\LTcapwidth}{\textwidth}
\begin{longtable}[c]{@{}>{\raggedright\arraybackslash}p{1.85cm}l l l>{\raggedright\arraybackslash}p{1.85cm}>{\raggedright\arraybackslash}p{8.10cm}@{}}
\hline
\textbf{Category} & \textbf{Log fpm} & \textbf{Band} & \textbf{ZH} & \textbf{English} & \textbf{Accepted English forms} \\
\hline
\endfirsthead
\hline
\textbf{Category} & \textbf{Log fpm} & \textbf{Band} & \textbf{ZH} & \textbf{English} & \textbf{Accepted English forms} \\
\hline
\endhead
\endfoot
\endlastfoot
Animal & 1.347 & high & 兔子 & rabbit & rabbit; rabbits; bunny; bunnies; hare; hares \\
Animal & 1.336 & high & 蚊子 & mosquito & mosquito; mosquitoes; mosquitos \\
Animal & 1.275 & high & 猴子 & monkey & monkey; monkeys \\
Animal & 1.238 & high & 老虎 & tiger & tiger; tigers \\
Animal & 1.183 & high & 蝴蝶 & butterfly & butterfly; butterflies \\
Animal & 1.180 & high & 蜜蜂 & bee & bee; bees; honeybee; honeybees; honey bee; honey bees \\
Animal & 1.154 & high & 狮子 & lion & lion; lions \\
Animal & 1.149 & high & 熊猫 & panda & panda; pandas; giant panda; giant pandas \\
Animal & 1.078 & high & 蚂蚁 & ant & ant; ants \\
Animal & 1.069 & high & 苍蝇 & fly & fly; flies; housefly; houseflies; house fly; house flies \\
Animal & 0.861 & medium & 天鹅 & swan & swan; swans \\
Animal & 0.818 & medium & 孔雀 & peacock & peacock; peacocks; peafowl; peafowls \\
Animal & 0.814 & medium & 公鸡 & rooster & rooster; roosters; cockerel; cockerels; cock; cocks \\
Animal & 0.805 & medium & 麻雀 & sparrow & sparrow; sparrows \\
Animal & 0.785 & medium & 燕子 & swallow & swallow; swallows \\
Animal & 0.770 & medium & 松鼠 & squirrel & squirrel; squirrels \\
Animal & 0.704 & medium & 鹦鹉 & parrot & parrot; parrots \\
Animal & 0.669 & medium & 鲨鱼 & shark & shark; sharks \\
Animal & 0.598 & medium & 鳄鱼 & crocodile & crocodile; crocodiles; alligator; alligators \\
Animal & 0.595 & medium & 喜鹊 & magpie & magpie; magpies \\
Animal & 0.375 & low & 水母 & jellyfish & jellyfish; jellyfishes \\
Animal & 0.374 & low & 海龟 & sea turtle & sea turtle; sea turtles; marine turtle; marine turtles \\
Animal & 0.343 & low & 斑马 & zebra & zebra; zebras \\
Animal & 0.277 & low & 海豹 & seal & seal; seals \\
Animal & 0.276 & low & 犀牛 & rhinoceros & rhinoceros; rhinoceroses; rhino; rhinos \\
Animal & 0.253 & low & 河马 & hippopota-\newline mus & hippopotamus; hippopotamuses; hippopotami; hippo; hippos \\
Animal & 0.170 & low & 壁虎 & gecko & gecko; geckos; geckoes \\
Animal & 0.150 & low & 海狮 & sea lion & sea lion; sea lions \\
Animal & 0.116 & low & 蟒蛇 & python & python; pythons; boa; boas; boa constrictor; boa constrictors \\
Animal & 0.112 & low & 乌贼 & cuttlefish & cuttlefish; cuttlefishes; cuttle fish; cuttle fishes; squid; squids \\
Fruit & 1.658 & high & 苹果 & apple & apple; apples \\
Fruit & 1.396 & high & 西瓜 & watermelon & watermelon; watermelons \\
Fruit & 1.288 & high & 葡萄 & grape & grape; grapes \\
Fruit & 1.246 & high & 榴莲 & durian & durian; durians \\
Fruit & 1.140 & high & 芒果 & mango & mango; mangoes; mangos \\
Fruit & 1.115 & high & 香蕉 & banana & banana; bananas \\
Fruit & 1.040 & high & 樱桃 & cherry & cherry; cherries \\
Fruit & 1.008 & high & 橘子 & mandarin & mandarin; mandarins; mandarin orange; mandarin oranges; tangerine; tangerines; orange; oranges \\
Fruit & 0.956 & high & 桃子 & peach & peach; peaches \\
Fruit & 0.850 & high & 柠檬 & lemon & lemon; lemons \\
Fruit & 0.482 & medium & 梅子 & plum & plum; plums; chinese plum; chinese plums; japanese apricot; japanese apricots \\
Fruit & 0.447 & medium & 鸭梨 & ya pear & ya pear; ya pears; yali pear; yali pears; chinese pear; chinese pears; pear; pears \\
Fruit & 0.414 & medium & 甜瓜 & melon & melon; melons; sweet melon; sweet melons; muskmelon; muskmelons; cantaloupe; cantaloupes \\
Fruit & 0.407 & medium & 龙眼 & longan & longan; longans; longan fruit; longan fruits; dragon eye; dragon eyes; dragon's eye; dragon's eyes \\
Fruit & 0.405 & medium & 橄榄 & olive & olive; olives \\
Fruit & 0.385 & medium & 蓝莓 & blueberry & blueberry; blueberries \\
Fruit & 0.300 & medium & 黑莓 & blackberry & blackberry; blackberries \\
Fruit & 0.295 & medium & 梨子 & pear & pear; pears \\
Fruit & 0.290 & medium & 杏子 & apricot & apricot; apricots \\
Fruit & 0.254 & medium & 桑葚 & mulberry & mulberry; mulberries \\
Fruit & 0.078 & low & 释迦 & sugar apple & sugar apple; sugar apples; sweetsop; sweetsops; custard apple; custard apples \\
Fruit & 0.073 & low & 椰枣 & date & date; dates; date fruit; date fruits \\
Fruit & 0.072 & low & 芭乐 & guava & guava; guavas \\
Fruit & 0.069 & low & 香梨 & fragrant pear & fragrant pear; fragrant pears; korla pear; korla pears; chinese pear; chinese pears; pear; pears \\
Fruit & 0.054 & low & 青枣 & green date & green date; green dates; green jujube; green jujubes; jujube; jujubes; chinese date; chinese dates; indian jujube; indian jujubes \\
Fruit & 0.036 & low & 莲雾 & wax apple & wax apple; wax apples; java apple; java apples; water apple; water apples; rose apple; rose apples; wax jambu; wax jambus \\
Fruit & 0.035 & low & 沙果 & crabapple & crabapple; crabapples; crab apple; crab apples; chinese crabapple; chinese crabapples \\
Fruit & 0.033 & low & 蛇果 & red delicious apple & red delicious apple; red delicious apples; red delicious; apple; apples \\
Fruit & 0.031 & low & 红提 & red grape & red grape; red grapes; red seedless grape; red seedless grapes; grape; grapes \\
Fruit & 0.020 & low & 血橙 & blood orange & blood orange; blood oranges; orange; oranges \\
Music & 1.172 & high & 钢琴 & piano & piano; pianos \\
Music & 1.127 & high & 小号 & trumpet & trumpet; trumpets \\
Music & 0.932 & high & 吉他 & guitar & guitar; guitars \\
Music & 0.739 & high & 大号 & tuba & tuba; tubas \\
Music & 0.730 & high & 号角 & horn & horn; horns; bugle; bugles \\
Music & 0.666 & high & 琵琶 & pipa & pipa; pipas; chinese lute; chinese lutes \\
Music & 0.588 & high & 锣鼓 & percussion & percussion; percussion instruments; gongs and drums; gong and drum \\
Music & 0.511 & high & 铃铛 & bell & bell; bells \\
Music & 0.483 & high & 笛子 & flute & flute; flutes \\
Music & 0.445 & high & 大鼓 & bass drum & bass drum; bass drums; big drum; big drums; large drum; large drums \\
Music & 0.187 & medium & 腰鼓 & waist drum & waist drum; waist drums \\
Music & 0.172 & medium & 铜鼓 & bronze drum & bronze drum; bronze drums \\
Music & 0.140 & medium & 风笛 & bagpipe & bagpipe; bagpipes \\
Music & 0.128 & medium & 花鼓 & flower drum & flower drum; flower drums; flower-drum; flower-drums; huagu drum; huagu drums \\
Music & 0.123 & medium & 长号 & trombone & trombone; trombones \\
Music & 0.120 & medium & 竹板 & bamboo clapper & bamboo clapper; bamboo clappers; bamboo clapperboard; bamboo clapperboards; kuaiban \\
Music & 0.111 & medium & 贝斯 & bass & bass; basses; bass guitar; bass guitars \\
Music & 0.110 & medium & 手鼓 & hand drum & hand drum; hand drums; tambourine; tambourines \\
Music & 0.108 & medium & 芦笙 & lusheng & lusheng; lushengs; reed pipe; reed pipes; reed-pipe; reed-pipes \\
Music & 0.095 & medium & 短笛 & piccolo & piccolo; piccolos \\
Music & 0.015 & low & 柳琴 & liuqin & liuqin; liuqins; liu qin; chinese mandolin; chinese mandolins \\
Music & 0.014 & low & 编磬 & stone chimes & stone chimes; stone chime; lithophone; lithophones; qing \\
Music & 0.012 & low & 高胡 & gaohu & gaohu; gaohus; gao hu; cantonese fiddle; cantonese fiddles \\
Music & 0.012 & low & 边鼓 & side drum & side drum; side drums; snare drum; snare drums \\
Music & 0.011 & low & 手铃 & handbell & handbell; handbells; hand bell; hand bells \\
Music & 0.011 & low & 沙锤 & maracas & maracas; maraca; shaker; shakers \\
Music & 0.010 & low & 钢鼓 & steel drum & steel drum; steel drums; steelpan; steelpans; steel pan; steel pans \\
Music & 0.009 & low & 陶笛 & ocarina & ocarina; ocarinas \\
Music & 0.007 & low & 钟琴 & glocken-\newline spiel & glockenspiel; glockenspiels; orchestral bells; bell lyre; bell lyres \\
Music & 0.006 & low & 坠胡 & zhuihu & zhuihu; zhuihus; zhui hu; zhuiqin; zhui qin \\
Occupation & 2.571 & high & 记者 & reporter & reporter; reporters; journalist; journalists; correspondent; correspondents \\
Occupation & 2.350 & high & 农民 & farmer & farmer; farmers; peasant; peasants; agricultural worker; agricultural workers \\
Occupation & 2.290 & high & 教师 & teacher & teacher; teachers; instructor; instructors; educator; educators \\
Occupation & 2.226 & high & 工人 & worker & worker; workers; laborer; laborers; labourer; labourers \\
Occupation & 2.087 & high & 医生 & doctor & doctor; doctors; physician; physicians; medical doctor; medical doctors \\
Occupation & 1.882 & high & 教授 & professor & professor; professors \\
Occupation & 1.827 & high & 学者 & scholar & scholar; scholars; academic; academics; researcher; researchers \\
Occupation & 1.817 & high & 作家 & writer & writer; writers; author; authors \\
Occupation & 1.773 & high & 警察 & police officer & police officer; police officers; police; policeman; policemen; policewoman; policewomen; cop; cops \\
Occupation & 1.735 & high & 演员 & actor & actor; actors; actress; actresses; performer; performers \\
Occupation & 1.342 & medium & 导演 & director & director; directors; film director; film directors; movie director; movie directors \\
Occupation & 1.321 & medium & 画家 & painter & painter; painters; artist; artists \\
Occupation & 1.250 & medium & 商人 & merchant & merchant; merchants; trader; traders; businessman; businessmen; businesswoman; businesswomen; businessperson; businesspeople \\
Occupation & 1.215 & medium & 歌手 & singer & singer; singers; vocalist; vocalists \\
Occupation & 1.196 & medium & 船长 & captain & captain; captains; ship captain; ship captains \\
Occupation & 1.164 & medium & 助理 & assistant & assistant; assistants; aide; aides \\
Occupation & 1.133 & medium & 裁判 & referee & referee; referees; umpire; umpires; judge; judges \\
Occupation & 1.132 & medium & 模特 & model & model; models; fashion model; fashion models \\
Occupation & 1.109 & medium & 牧师 & pastor & pastor; pastors; minister; ministers; priest; priests \\
Occupation & 1.069 & medium & 保安 & security guard & security guard; security guards; security officer; security officers; guard; guards \\
Occupation & 0.619 & low & 电工 & electrician & electrician; electricians \\
Occupation & 0.538 & low & 园丁 & gardener & gardener; gardeners; groundskeeper; groundskeepers \\
Occupation & 0.419 & low & 画师 & illustrator & illustrator; illustrators; artist; artists; painter; painters \\
Occupation & 0.346 & low & 牙医 & dentist & dentist; dentists \\
Occupation & 0.297 & low & 舞者 & dancer & dancer; dancers \\
Occupation & 0.268 & low & 花匠 & florist & florist; florists; gardener; gardeners; flower grower; flower growers \\
Occupation & 0.246 & low & 茶农 & tea farmer & tea farmer; tea farmers; tea grower; tea growers; tea planter; tea planters \\
Occupation & 0.166 & low & 乐手 & musician & musician; musicians; instrumentalist; instrumentalists \\
Occupation & 0.064 & low & 码农 & program-\newline mer & programmer; programmers; coder; coders; software developer; software developers; software engineer; software engineers; developer; developers \\
Occupation & 0.038 & low & 机师 & pilot & pilot; pilots; aviator; aviators; airline pilot; airline pilots; aircraft pilot; aircraft pilots \\
Vehicle & 2.124 & high & 汽车 & car & car; cars; automobile; automobiles; auto; autos; motorcar; motorcars \\
Vehicle & 1.971 & high & 飞机 & airplane & airplane; airplanes; aeroplane; aeroplanes; plane; planes; aircraft \\
Vehicle & 1.708 & high & 火车 & train & train; trains \\
Vehicle & 1.503 & high & 地铁 & subway & subway; subways; metro; metros; underground; underground train; underground trains; tube \\
Vehicle & 1.479 & high & 马车 & carriage & carriage; carriages; horse-drawn carriage; horse-drawn carriages; horse carriage; horse carriages; horse cart; horse carts; wagon; wagons \\
Vehicle & 1.349 & high & 列车 & train & train; trains; railway train; railway trains \\
Vehicle & 1.245 & high & 高铁 & high-speed train & high-speed train; high-speed trains; high speed train; high speed trains; bullet train; bullet trains; hsr \\
Vehicle & 1.157 & high & 轿车 & sedan & sedan; sedans; saloon; saloons; car; cars \\
Vehicle & 1.110 & high & 卡车 & truck & truck; trucks; lorry; lorries \\
Vehicle & 1.060 & high & 火箭 & rocket & rocket; rockets \\
Vehicle & 0.611 & medium & 巴士 & bus & bus; buses; busses; coach; coaches \\
Vehicle & 0.605 & medium & 警车 & police car & police car; police cars; patrol car; patrol cars \\
Vehicle & 0.564 & medium & 游艇 & yacht & yacht; yachts; pleasure boat; pleasure boats \\
Vehicle & 0.563 & medium & 潜艇 & submarine & submarine; submarines; sub; subs \\
Vehicle & 0.562 & medium & 滑板 & skateboard & skateboard; skateboards \\
Vehicle & 0.530 & medium & 战机 & fighter jet & fighter jet; fighter jets; fighter plane; fighter planes; fighter aircraft; warplane; warplanes; fighter; fighters \\
Vehicle & 0.507 & medium & 战车 & chariot & chariot; chariots; war chariot; war chariots; combat vehicle; combat vehicles; military vehicle; military vehicles; tank; tanks \\
Vehicle & 0.503 & medium & 越野 & off-road vehicle & off-road vehicle; off-road vehicles; off road vehicle; off road vehicles; off-roader; off-roaders; 4x4; suv; suvs; jeep; jeeps \\
Vehicle & 0.500 & medium & 皮卡 & pickup truck & pickup truck; pickup trucks; pickup; pickups; pick-up truck; pick-up trucks; pick-up; pick-ups \\
Vehicle & 0.487 & medium & 校车 & school bus & school bus; school buses; school busses \\
Vehicle & 0.228 & low & 客船 & passenger ship & passenger ship; passenger ships; passenger boat; passenger boats; passenger ferry; passenger ferries \\
Vehicle & 0.194 & low & 罐车 & tanker & tanker; tankers; tank truck; tank trucks; tanker truck; tanker trucks; tank car; tank cars \\
Vehicle & 0.190 & low & 飞艇 & airship & airship; airships; dirigible; dirigibles; blimp; blimps \\
Vehicle & 0.167 & low & 叉车 & forklift & forklift; forklifts; fork-lift; fork-lifts; lift truck; lift trucks \\
Vehicle & 0.142 & low & 竹筏 & bamboo raft & bamboo raft; bamboo rafts; raft; rafts \\
Vehicle & 0.138 & low & 渡轮 & ferry & ferry; ferries; ferryboat; ferryboats; ferry boat; ferry boats \\
Vehicle & 0.117 & low & 房车 & recreational vehicle & recreational vehicle; recreational vehicles; rv; rvs; motorhome; motorhomes; motor home; motor homes; camper van; camper vans; caravan; caravans \\
Vehicle & 0.075 & low & 兵车 & military vehicle & military vehicle; military vehicles; army vehicle; army vehicles; troop carrier; troop carriers; war vehicle; war vehicles \\
Vehicle & 0.044 & low & 驴车 & donkey cart & donkey cart; donkey carts; donkey-drawn cart; donkey-drawn carts; donkey carriage; donkey carriages \\
Vehicle & 0.027 & low & 铲车 & loader & loader; loaders; front loader; front loaders; wheel loader; wheel loaders; front-end loader; front-end loaders \\
\hline
\caption{Complete lexical inventory and accepted English forms. ZH denotes the Mandarin word; Log fpm is $\log_{10}(\mathrm{fpm}+1)$, where fpm is frequency per million BCC tokens. Within each category, the 30 retained words are listed in descending order of frequency.}
\label{tab:complete-lexical-inventory}\\
\end{longtable}
}

\twocolumn

\raggedbottom
\section{Complete Prompt Materials}
\label{app:prompts}

\subsection{Shared Prompt Components}
\label{app:prompt-shell}

The boxes below give the shared instruction, prompt frames, sentence templates, and illustrative examples. Bracketed fields mark slots filled during prompt generation. Underlining links each Mandarin word to its familiar English translation in the reader annotations. Underlining and parenthetical English translations were not included in model input.

\begin{promptbox}{Shared instruction and prompt shell}
\texttt{Task:} Translate Mandarin to English. Output only the English translation without explanations or notes.

\texttt{[FRAME]}

\texttt{Translate: [SOURCE SENTENCE]}
\end{promptbox}

\begin{promptbox}{Prompt-frame templates}
\textbf{F1:} Use what you know about Mandarin and ignore any newly created lexical rule. \texttt{[SOURCE]} means \texttt{[TARGET]}.

\textbf{F2:} Lexicon: \texttt{[SOURCE]} means \texttt{[TARGET]}.

\textbf{F3:} We are using a newly documented variety of Mandarin. In this variety, \texttt{[SOURCE]} means \texttt{[TARGET]}.

\textbf{F4:} We are playing a language game. In this game, \texttt{[SOURCE]} means \texttt{[TARGET]}.

\textbf{F5:} Use the lexicon below even if it conflicts with standard Mandarin. \texttt{[SOURCE]} means \texttt{[TARGET]}.
\end{promptbox}

\begin{promptbox}{Source-sentence templates}
\textbf{Animal:} 同学们在生物课上学了\texttt{[SOURCE]}的习性。

\textit{(Translation: The students learned about the habits of [SOURCE] in biology class.)}

\textbf{Fruit:} 这些\texttt{[SOURCE]}已经熟到可以吃了。

\textit{(Translation: These [SOURCE] are ripe enough to eat.)}

\textbf{Vehicle:} 我们坐\texttt{[SOURCE]}抵达了目的地。

\textit{(Translation: We arrived at the destination by [SOURCE].)}

\textbf{Occupation:} 我在毕业多年以后成为了\texttt{[SOURCE]}。

\textit{(Translation: I became a [SOURCE] many years after graduating.)}

\textbf{Music:} 那位音乐家在舞台上演奏着\texttt{[SOURCE]}。

\textit{(Translation: The musician was playing [SOURCE] on stage.)}
\end{promptbox}

\subsection{Baseline Prompts}
\label{app:baseline-prompts}

Both no-rule baselines omit the lexical rule. The semantically consistent baseline inserts a word into its own category's sentence template; the semantically inconsistent baseline uses another category's template.

\begin{promptbox}{Semantically consistent baseline}
\texttt{Task:} Translate Mandarin to English. Output only the English translation without explanations or notes.

\texttt{Translate:} 同学们在生物课上学了\underline{蝴蝶}的习性。

\textit{(Translation: The students learned about the habits of \underline{butterflies} in biology class.)}
\end{promptbox}

\begin{promptbox}{Semantically inconsistent baseline}
\texttt{Task:} Translate Mandarin to English. Output only the English translation without explanations or notes.

\texttt{Translate:} 这些\underline{蝴蝶}已经熟到可以吃了。

\textit{(Translation: These \underline{butterflies} are ripe enough to eat.)}
\end{promptbox}

Each item appears once in its matching category and once in each of the four nonmatching categories, yielding 750 baseline prompts per model.

\subsection{Experiment 1: Lexical Conflicts across Semantic Categories}
\label{app:exp1-prompts}

Each mapping occurs under all five frames and both sentence contexts. For illustration, the five boxes instantiate different mappings in the category cycle Animal$\rightarrow$Fruit, Fruit$\rightarrow$Vehicle, Vehicle$\rightarrow$Occupation, Occupation$\rightarrow$Music, and Music$\rightarrow$Animal. The first two examples use PK contexts and the remaining three use CK contexts. This allocation is for presentation only: both contexts occur under every frame in the experiment.

\begin{promptbox}{Experiment 1, F1: PK context}
\texttt{Task:} Translate Mandarin to English. Output only the English translation without explanations or notes.

Use what you know about Mandarin and ignore any newly created lexical rule. \underline{蝴蝶}\ means red grape.

\texttt{Translate:} 同学们在生物课上学了\underline{蝴蝶}的习性。

\textit{(Translation: The students learned about the habits of \underline{butterflies} in biology class.)}
\end{promptbox}

\begin{promptbox}{Experiment 1, F2: PK context}
\texttt{Task:} Translate Mandarin to English. Output only the English translation without explanations or notes.

Lexicon: \underline{苹果}\ means bus.

\texttt{Translate:} 这些\underline{苹果}已经熟到可以吃了。

\textit{(Translation: These \underline{apples} are ripe enough to eat.)}
\end{promptbox}

\begin{promptbox}{Experiment 1, F3: CK context}
\texttt{Task:} Translate Mandarin to English. Output only the English translation without explanations or notes.

We are using a newly documented variety of Mandarin. In this variety, \underline{汽车}\ means programmer.

\texttt{Translate:} 我在毕业多年以后成为了\underline{汽车}。

\textit{(Translation: I became a \underline{car} many years after graduating.)}
\end{promptbox}

\begin{promptbox}{Experiment 1, F4: CK context}
\texttt{Task:} Translate Mandarin to English. Output only the English translation without explanations or notes.

We are playing a language game. In this game, \underline{码农}\ means piano.

\texttt{Translate:} 那位音乐家在舞台上演奏着\underline{码农}。

\textit{(Translation: The musician was playing a \underline{programmer} on stage.)}
\end{promptbox}

\begin{promptbox}{Experiment 1, F5: CK context}
\texttt{Task:} Translate Mandarin to English. Output only the English translation without explanations or notes.

Use the lexicon below even if it conflicts with standard Mandarin. \underline{钢琴}\ means rabbit.

\texttt{Translate:} 同学们在生物课上学了\underline{钢琴}的习性。

\textit{(Translation: The students learned about the habits of \underline{pianos} in biology class.)}
\end{promptbox}

\subsection{Experiment 2: Lexical Conflicts within Semantic Categories}
\label{app:exp2-prompts}

Experiment~2 uses one within-category sentence per mapping. The five examples cover Animal, Fruit, Vehicle, Occupation, and Music in F1--F5 order. Each pair draws its source and target items from different frequency bands.

\begin{promptbox}{Experiment 2, F1: Animal}
\texttt{Task:} Translate Mandarin to English. Output only the English translation without explanations or notes.

Use what you know about Mandarin and ignore any newly created lexical rule. \underline{兔子}\ means crocodile.

\texttt{Translate:} 同学们在生物课上学了\underline{兔子}的习性。

\textit{(Translation: The students learned about the habits of \underline{rabbits} in biology class.)}
\end{promptbox}

\begin{promptbox}{Experiment 2, F2: Fruit}
\texttt{Task:} Translate Mandarin to English. Output only the English translation without explanations or notes.

Lexicon: \underline{苹果}\ means red grape.

\texttt{Translate:} 这些\underline{苹果}已经熟到可以吃了。

\textit{(Translation: These \underline{apples} are ripe enough to eat.)}
\end{promptbox}

\begin{promptbox}{Experiment 2, F3: Vehicle}
\texttt{Task:} Translate Mandarin to English. Output only the English translation without explanations or notes.

We are using a newly documented variety of Mandarin. In this variety, \underline{汽车}\ means bus.

\texttt{Translate:} 我们坐\underline{汽车}抵达了目的地。

\textit{(Translation: We arrived at the destination by \underline{car}.)}
\end{promptbox}

\begin{promptbox}{Experiment 2, F4: Occupation}
\texttt{Task:} Translate Mandarin to English. Output only the English translation without explanations or notes.

We are playing a language game. In this game, \underline{码农}\ means doctor.

\texttt{Translate:} 我在毕业多年以后成为了\underline{码农}。

\textit{(Translation: I became a \underline{programmer} many years after graduating.)}
\end{promptbox}

\begin{promptbox}{Experiment 2, F5: Music}
\texttt{Task:} Translate Mandarin to English. Output only the English translation without explanations or notes.

Use the lexicon below even if it conflicts with standard Mandarin. \underline{钢琴}\ means ocarina.

\texttt{Translate:} 那位音乐家在舞台上演奏着\underline{钢琴}。

\textit{(Translation: The musician was playing \underline{piano} on stage.)}
\end{promptbox}

\subsection{Controlled Follow-up Prompts}
\label{app:followup-prompts}

\paragraph{Modal weakening.}
Modal weakening replaces \emph{means} with \emph{may mean} in each definition. The frame and translation request remain the same as in Experiment~1.

\begin{promptbox}{Modal weakening, F1: PK context}
\texttt{Task:} Translate Mandarin to English. Output only the English translation without explanations or notes.

Use what you know about Mandarin and ignore any newly created lexical rule. \underline{蝴蝶}\ may mean red grape.

\texttt{Translate:} 同学们在生物课上学了\underline{蝴蝶}的习性。

\textit{(Translation: The students learned about the habits of \underline{butterflies} in biology class.)}
\end{promptbox}

% ===== Begin inline appendix: appendix-position-repetition-prompts =====
\paragraph{Lexical-rule reiteration and position.}
\label{app:reiteration-prompts}
B denotes a position before the task instruction, M the position between that instruction and the source sentence, and E a position after the source sentence. The task instruction and translation request each occur once, in that order. The ten layouts are:

\begin{center}
\setlength{\tabcolsep}{2.5pt}
\begin{tabular}{@{}lcl@{}}
\hline
\textbf{Layout} & \textbf{Copies} & \textbf{Occupied positions} \\
\hline
B & 1 & Before task \\
M & 1 & Middle \\
E & 1 & After source \\
M$\times$2 & 2 & Middle, consecutive \\
BM & 2 & Before task; middle \\
BE & 2 & Before task; after source \\
ME & 2 & Middle; after source \\
M$\times$3 & 3 & Middle, consecutive \\
BME & 3 & Before task; middle; after source \\
M$\times$5 & 5 & Middle, consecutive \\
\hline
\end{tabular}
\end{center}

The first copy of the rule always contains the complete frame instruction, followed by the lexical rule. Any subsequent copies repeat only the rule. The following examples use the same lexical rule, F5, and a familiar-compatible sentence.

\begin{promptbox}{Two middle copies (M$\times$2), F5}
\texttt{Task:} Translate Mandarin to English. Output only the English translation without explanations or notes.

Use the lexicon below even if it conflicts with standard Mandarin. \underline{兔子}\ means green date. \underline{兔子}\ means green date.

\texttt{Translate:} 同学们在生物课上学了\underline{兔子}的习性。

\textit{(Translation: The students learned about the habits of \underline{rabbits} in biology class.)}
\end{promptbox}

The five-copy middle layout uses the same construction, with five copies of the lexical rule:

\begin{promptbox}{Five middle copies (M$\times$5), F5}
\texttt{Task:} Translate Mandarin to English. Output only the English translation without explanations or notes.

Use the lexicon below even if it conflicts with standard Mandarin. \underline{兔子}\ means green date. \underline{兔子}\ means green date. \underline{兔子}\ means green date. \underline{兔子}\ means green date. \underline{兔子}\ means green date.

\texttt{Translate:} 同学们在生物课上学了\underline{兔子}的习性。

\textit{(Translation: The students learned about the habits of \underline{rabbits} in biology class.)}
\end{promptbox}

Two distributed copies, one before the task (B) and one after the source (E), elicited the highest CK rate of all layouts:

\begin{promptbox}{Before task and after source (BE), F5}
Use the lexicon below even if it conflicts with standard Mandarin. \underline{兔子}\ means green date.

\texttt{Task:} Translate Mandarin to English. Output only the English translation without explanations or notes.

\texttt{Translate:} 同学们在生物课上学了\underline{兔子}的习性。

\underline{兔子}\ means green date.

\textit{(Translation: The students learned about the habits of \underline{rabbits} in biology class.)}
\end{promptbox}

Full results for all nine models appear in Appendix~\ref{app:paper-position-repetition}.
% ===== End inline appendix: appendix-position-repetition-prompts =====

\paragraph{Explicit reasoning.}
The reasoning follow-up uses the Experiment~1 prompts verbatim, with medium reasoning enabled for GPT-5.5 and thinking mode enabled for Qwen3-8B and Qwen3-32B. Appendix~\ref{app:inference-details} gives the output budgets and recovery settings.

% Finish Appendix B with natural spacing before restoring flush-bottom columns.
\clearpage
\flushbottom
\section{Experimental Construction, Analysis Specifications, and Audits}
\label{app:construction}

\subsection{Experiment 1 Source--Target Mapping}
\label{app:exp1-construction}

Experiment~1 includes all 20 directed pairs among the five semantic categories, each with 30 source--target mappings (600 in total) in which every source and target item appears once. Pairings used global seed 2026 plus the directed-pair index. Crossing the mappings with five frames and two contexts yields 6,000 prompts per model.

\label{app:frequency-balancing}
Within each directed pair, high-, medium-, and low-frequency sources are assigned to high-, medium-, and low-frequency targets in 4/3/3, 3/4/3, and 3/3/4 splits, respectively. This keeps ten sources and ten targets per frequency band while avoiding deterministic rank-to-rank matching.

\subsection{Experiment 2 Directed Mappings and Randomization}
\label{app:exp2-construction}

Experiment~2 pairs items within semantic categories, pairing each source band with each of the other two target bands (no same-band mappings). For each ordered band pair, ten sources are assigned one-to-one to ten targets in two independently randomized rounds (seeds 2027 and 2028), giving $5 \times 3 \times 2 \times 10 \times 2 = 600$ directed mappings and, with five frames and one within-category context, 3,000 prompts per model.

\label{app:overlap-adjustment}
An audit of the initial mappings flagged 33 rows: 28 ordered source--target pairs repeated across rounds and five paired items sharing an accepted English form. Seventeen local swaps resolved all flags; each exchanged targets between rows adjacent in source frequency rank within the same category, source band, target band, and round, preserving the balancing constraints. The final design contains no repeated ordered pairs and no shared complete forms.

\begin{table*}[t]
  \centering
  \small
  \setlength{\tabcolsep}{5pt}
  \begin{tabular}{@{}>{\raggedright\arraybackslash}p{2.90cm}>{\raggedright\arraybackslash}p{3.60cm}>{\raggedright\arraybackslash}p{2.10cm}>{\raggedright\arraybackslash}p{2.40cm}>{\raggedright\arraybackslash}p{3.00cm}@{}}
    \hline
    \textbf{Model} & \textbf{Runtime identifier} & \textbf{Size} & \textbf{Interface} & \textbf{Reasoning setting} \\
    \hline
    GPT-5.5 & \texttt{gpt-5.5} & Undisclosed & OpenAI API & \texttt{none} (Thinking off) \\
    Qwen3-8B & \texttt{qwen3:8b} & 8B & Ollama & Thinking off \\
    Qwen3-32B & \texttt{qwen3:32b} & 32B & Ollama & Thinking off \\
    Llama-3-8B & \texttt{llama3:8b} & 8B & Ollama & No thinking switch \\
    Llama-3-70B & \texttt{llama3:70b} & 70B & Ollama & No thinking switch \\
    DeepSeek-LLM-7B & \texttt{deepseek-llm:7b} & 7B & Ollama & No thinking switch \\
    DeepSeek-LLM-67B & \texttt{deepseek-llm:67b} & 67B & Ollama & No thinking switch \\
    Gemma-3-4B & \texttt{gemma3:4b} & 4B & Ollama & No thinking switch \\
    Gemma-3-27B & \texttt{gemma3:27b} & 27B & Ollama & No thinking switch \\
    \hline
  \end{tabular}
  \caption{Model identities and inference interfaces. Runtime identifiers reproduce the exact model strings stored with the responses.}
  \label{tab:model-runtime-identifiers}
\end{table*}

\subsection{Model and Inference Configuration}
\label{app:inference-details}

Table~\ref{tab:model-runtime-identifiers} lists the evaluated models; sizes are nominal, and OpenAI does not disclose GPT-5.5's parameter count. The Ollama runner (v0.13.1; \citealp{ollama2025}) queried the native \texttt{/api/generate} endpoint with streaming disabled; the OpenAI runner queried \texttt{/v1/responses} with response storage disabled. GPU-based experiments used NVIDIA A100 GPUs, with an estimated total allocation of approximately 1,200 GPU-hours.

Standard Ollama requests set \texttt{temperature=0}, \texttt{seed=2026}, \texttt{num\_ctx=4096}, and \texttt{num\_predict=128}, leaving other options at runtime defaults; Qwen3 thinking was disabled. GPT-5.5 requests set \texttt{max\_output\_tokens=128}, \texttt{reasoning.effort=none}, and \texttt{text.verbosity=low}, without temperature or seed. In the original study, a response that did not complete within this allowance received a targeted recovery request, and we kept the latest complete, non-empty response. The eight additional lexical-rule layouts (Appendix~\ref{app:paper-position-repetition}) used the standard settings without recovery; seven of their 432,000 responses were incomplete and are retained as \textsc{Failed}.

The final files contain 175 recoveries in Experiment~1, 177 in Experiment~2, 129 in the modal-weakening follow-up, and 66 in the reiteration follow-up; most only raised the output ceiling to 1,048 tokens. Seven Llama-3-8B reiteration recoveries were adjusted individually: three used a 1,048-token ceiling and four an 8,192-token context with a 4,096-token ceiling (three of these also with seed 2027 and nonzero temperature). Exact settings are stored with every response row.

In the explicit-reasoning runs, Qwen3 used thinking mode with a 1,048-token ceiling (11,731 responses) or an 8,192-token context and 4,096-token ceiling (269). GPT-5.5 used medium reasoning with the standard 128-token ceiling (5,310) or a 1,048-token ceiling (690 recoveries). Prompts and conditions were unchanged.

\subsection{Analysis Specifications and Design Audits}
\label{app:analysis-audits}

The Experiment~1 mixed model predicts \textsc{CK} as
\begin{equation}
\label{eq:exp1-glmm}
\begin{aligned}
&\text{CK} \sim \text{context} + \text{frame} + \text{source frequency} \\
&\quad{}+ \text{target-frequency proxy} + (1 \mid \text{model}) \\
&\quad{}+ (1 \mid \text{source category}) + (1 \mid \text{mapping}) \\
&\quad{}+ (1 \mid \text{source item}) + (1 \mid \text{target item}) .
\end{aligned}
\end{equation}
The Experiment~2 model omits context, as each mapping has one within-category sentence. Separate source- and target-item intercepts account for each item's reuse across mappings and roles. Both specifications are additive with random intercepts only; treatment references are PK context (Experiment~1), F1, and the high bands of source frequency and the target-frequency proxy. We fit binomial logit models in \texttt{lme4} \citep{bates2015lme4} with \texttt{bobyqa} (at most 200,000 evaluations; \texttt{nAGQ=1}), testing joint effects with Wald tests and reporting unadjusted pointwise Wald 95\% confidence intervals. Both specifications were fit to the full (54,000 Experiment~1; 27,000 Experiment~2) and matched-baseline-filtered (42,270; 23,100) datasets; all four fits converged and passed the singularity check (tolerance $10^{-4}$).

Within each experiment and sample, Holm correction controls family-wise error at $\alpha=.05$ separately for omnibus fixed-effect tests (four in Experiment~1; three in Experiment~2) and for all non-intercept contrasts against the treatment references (nine and eight), giving eight families in total. Significance judgments use adjusted $p$ values; confidence intervals remain unadjusted.

Automated audits verified the expected numbers of items, mappings, prompt IDs, condition cells, and model runs, as well as unique identifiers and source--target constraints. The original study contains 6,750 no-rule baseline, 54,000 Experiment~1, 27,000 Experiment~2, 54,000 modal-weakening, 54,000 reiteration, and 18,000 explicit-reasoning responses (213,750 valid responses), with no missing or unresolved prompts. The eight additional lexical-rule layouts add 432,000 responses and the conventional-meaning baseline 6,750 (Appendices~\ref{app:paper-position-repetition} and~\ref{app:paper-conventional-meaning}).

\raggedbottom
\subsection{Choice of Random-Effects Structure}
\label{app:random-effects-choice}

We also tried two specifications with the same fixed effects that replace $(1 \mid \text{model})$ in Equation~\ref{eq:exp1-glmm} with model-specific coefficients (context terms in Experiment~1 only):
\begin{equation}
\label{eq:alternative-model-slopes}
\begin{aligned}
&(0 + \text{context} \mid \text{model}) \\
&\quad{}+ (0 + \text{frame} \mid \text{model}) \\
&\quad{}+ (0 + \text{source frequency} \mid \text{model}) .
\end{aligned}
\end{equation}
The first keeps the mapping, source-category, and item intercepts; the second drops them. Both produced singular fits in the full and baseline-filtered datasets, so we report Equation~\ref{eq:exp1-glmm} (omitting context in Experiment~2). The analysis repository provides the alternatives' fitting settings, convergence and singularity diagnostics, random-effect variances and correlations, and fixed-effect estimates.

\flushbottom
% ===== Begin inline appendix: appendix-model-results =====
% D/E prose stays in two columns; the main file emits the displays later.
\section{Full Model Results and Baseline-Filtered Robustness}
\label{app:filtered-robustness}

This appendix gives results for each model in the full and baseline-filtered datasets. Filtering retains only trials for which the same model produced a \textsc{Familiar} response on the matched no-rule baseline. Table~\ref{tab:filtered-coefficients} reports all non-intercept contrasts from the filtered GLMMs. Error bars in the figures show 95\% Wilson intervals \citep{wilson1927inference} for descriptive CK proportions.

\subsection{Experiment 1}
\label{app:exp1-full-results}

Figures~\ref{fig:exp1-unfiltered-full} and~\ref{fig:exp1-filtered} show response distributions and CK rates by condition for all nine models in the full and baseline-filtered datasets.

For PK-context trials, filtering required a \textsc{Familiar} response to the source item in the semantically consistent baseline; for CK-context trials, it required a \textsc{Familiar} response in the corresponding semantically inconsistent baseline. This retained 42,270 responses, of which 71.85\% were \textsc{CK}. Sentence context, prompt framing, and source frequency remained significant (context: $\chi^2(1)=3408.31$; frame: $\chi^2(4)=2927.09$; source frequency: $\chi^2(2)=38.45$; all $p<.001$), whereas the target-frequency proxy remained nonsignificant, $\chi^2(2)=1.57$, $p=.456$.

\subsection{Experiment 2}
\label{app:exp2-full-results}

Figures~\ref{fig:exp2-unfiltered-full} and~\ref{fig:exp2-filtered} show response distributions and CK rates by condition for all nine models in the full and baseline-filtered datasets.

In the descriptive comparison with Experiment~1, the pooled CK rate increased by 4.51 percentage points, with increases in all five frames (3.41--5.54 points). Eight models had higher CK rates; GPT-5.5 decreased from 84.32\% to 81.60\%. The larger model had the higher CK rate within every paired open-weight family in both experiments.

Baseline filtering required a \textsc{Familiar} response to the source item in its semantically consistent baseline. This retained 23,100 responses, with a CK rate of 77.80\%. Prompt framing and source frequency remained significant (frame: $\chi^2(4)=1859.13$; source frequency: $\chi^2(2)=18.30$; both $p<.001$), while the target-frequency proxy remained nonsignificant, $\chi^2(2)=3.29$, $p=.193$.

\section{Full Follow-up Results and Robustness Checks}
\label{app:followup-results}

Table~\ref{tab:followup-robustness} compares CK-rate changes in the full and baseline-filtered datasets. Each follow-up is compared with Experiment~1 for the same models and matched trials. The figures show response distributions and CK rates by condition, with 95\% Wilson intervals for descriptive CK proportions.

\subsection{Modal Weakening}
\label{app:modal-full-results}

Filtering retained 42,270 responses. The CK rate fell from 71.85\% in the matched Experiment~1 sample to 52.86\%, a decrease of 18.99 percentage points; the decrease in the full dataset was 17.73 points. Every model showed a decrease in both datasets (Figures~\ref{fig:modal-unfiltered-full} and~\ref{fig:modal-filtered-full}).

\subsection{Lexical-Rule Reiteration (M$\times$5)}
\label{app:reiteration-full-results}

Filtering retained 42,270 responses. Since the original follow-up experiment included only M$\times$5 and additional repetition and position experiments were conducted subsequently, we include detailed results for the M$\times$5 case here, and filtered results for all ten lexical-rule layouts appear in Appendix~\ref{app:paper-position-repetition}. The CK rate fell from 71.85\% in the matched Experiment~1 sample to 68.03\%, a decrease of 3.82 percentage points; the decrease in the full dataset was 3.42 points. The model-level directions were unchanged: Llama-3-8B and Qwen3-8B increased slightly, while the other seven models decreased (Figures~\ref{fig:reiteration-unfiltered-full} and~\ref{fig:reiteration-filtered-full}). 

\subsection{Explicit Reasoning and Thinking Mode}
\label{app:reasoning-full-results}

Filtering retained 15,440 responses across the three evaluated models. GPT-5.5's CK rate decreased and both Qwen models' rates increased, preserving the opposite directions observed in the full dataset (Figures~\ref{fig:reasoning-unfiltered-full} and~\ref{fig:reasoning-filtered-full}). The pooled CK rate rose from 75.27\% in the matched Experiment~1 sample to 92.14\%, a gain of 16.88 percentage points.

% ===== End inline appendix: appendix-model-results =====

\subsection{Translation Quality}
\label{app:translation-quality}

% ===== Begin inline appendix: appendix-translation-scores =====
% Source: ../results/translation-quality/scores/<condition>/<model>/literal_*_en.csv.
% Generated by extensions/paper-appendices/consistent-integration-20261010/scripts/make_translation_scores_appendix.py
% (means of the saved four-decimal sentence scores; chrF++ x100 to its native 0-100 scale). 10 October 2026.

In our study, we use a binary lexical inclusion score to systematically evaluate the extent to which models use novel CK translations in their outputs. Detailed evaluation of the effects of lexical conflict on translation quality is an open field of research beyond our study's scope. We therefore provide COMET \citep{rei-etal-2020-comet} and chrF++ \citep{popovic-2017-chrf} scores for each condition as a high-level supplement to support the findings in the main paper. COMET uses the \texttt{Unbabel/wmt22-comet-da} checkpoint, and sentence-level chrF++ is computed with SacreBLEU v2.6.0 using character order 6 and word order 2.

Because the meaning assigned to the target word depends on the prompt frame used in the experiment, we score each output with respect to both a \emph{Familiar} reference, which uses the source word's conventional (PK) meaning to translate the target word, and a \emph{Supplied} reference, which uses the prompt's assigned (CK) meaning to translate the target word. Both \emph{Familiar} and \emph{Supplied} references otherwise preserve the remaining sentence as an identical translation of the Mandarin source text, and all model outputs are scored verbatim against each reference.

Tables~\ref{tab:paper-metric-comet} and~\ref{tab:paper-metric-chrf} cover the no-rule and conventional-meaning baselines, both main experiments, modal weakening, all ten lexical-rule layouts, and the explicit-reasoning runs; the layouts are described in Appendix~\ref{app:reiteration-prompts}, and their CK rates appear in Appendix~\ref{app:paper-position-repetition}. COMET is reported on its native 0--1 scale and chrF++ on its 0--100 scale. Consistent with the results reported in the main paper, mean COMET and chrF++ scores are higher when measured against the \emph{Supplied} reference for every frame in both main experiments, including F1.

\paragraph{Coverage.}
In Experiment~1, the modal-weakening follow-up, and each lexical-rule layout, each frame contains 10,800 responses; 5,400 responses are evaluated in Experiment~2; 1,350 in the conventional-meaning baseline; and 3,600 in each reasoning-block row. The no-rule baselines contain 1,350 semantically consistent and 5,400 semantically inconsistent responses.

% Full-width tables are emitted after the D/E figures and statistics.

% ===== End inline appendix: appendix-translation-scores =====

% ===== Begin inline appendix: appendix-conventional-meaning-baseline =====
% Source: extensions/paper-appendices/consistent-integration-20261010/counts/conventional-original-inventory/counts.csv; original accepted-form inventory (data/coding/accepted_translations.csv) on both sides, as in Figure 2 and Appendix A.
\section{Additional Conventional-Meaning Baseline}
\label{app:paper-conventional-meaning}
 
We test an additional baseline that supplies
a lexical rule that agrees with the source word's familiar meaning. Each of the same 150
Mandarin source words appears in its own-category sentence under the five
unchanged frames. All nine models contribute 750 attempts each (150 words $\times$ five
frames), or 6,750 total. 
The matched no-rule comparison contains 150 historical own-category responses per model (1,350 unique responses). 

% The eight open-weight models used temperature 0, seed 2026, a 4,096-token context window and a 128-token output ceiling; Qwen3 thinking was disabled. GPT-5.5 used \texttt{gpt-5.5}, reasoning effort \texttt{none}, verbosity \texttt{low}, a 128-token output ceiling and disabled response storage, with no temperature or seed supplied. All 750 GPT-5.5 responses completed without retries. Unlike the recovered main-study runs, this appendix retains incomplete attempts and reports their coverage explicitly.

\paragraph{Lexical results.}
Automated matching searches the entire response; an accepted
form in an explanation also counts. Incomplete responses receive
\textsc{Failed} and do not count as lexical hits. Table~\ref{tab:paper-control-full}
reports familiar-form inclusion over all attempts, separately for every model
and frame.

Across models, familiar-form inclusion rises from 1,155/1,350 (85.56\%) without
a rule to 6,666/6,750 (98.76\%) with an agreeing rule. All 37 incomplete outputs
come from DeepSeek-LLM-67B and are nonempty responses truncated at 128 tokens.
Its completed-only rate is 707/713 (99.16\%); the pooled completed-only rate is
6,666/6,713 (99.30\%). All other models' denominators are unchanged.

\paragraph{Interpretation and limits.}
Agreeing lexical guidance is associated with near-ceiling inclusion of the
specified familiar rendering. A no-rule nonmatch can be an error or an
acceptable alternative outside the acceptance inventory; an explicit definition
can also disambiguate a sense or favor a particular rendering. 
Lexical inclusion therefore does not establish a corresponding gain in whole-sentence translation quality.

% ===== End inline appendix: appendix-conventional-meaning-baseline =====

% ===== Begin inline appendix: appendix-position-repetition =====
% Nine models, original accepted-form inventory and baseline screening labels.
% Counts: extensions/paper-appendices/consistent-integration-20261010/counts/ (code_layouts.py; tables by make_appendix_tables_v2.py).
\section{Full Lexical-Rule Reiteration and Position Results}
\label{app:paper-position-repetition}

This analysis compares ten lexical-rule layouts across all nine models. Each
layout uses the 600 cross-category mappings, five frames, and two sentence
contexts of Experiment~1: 6,000 prompts per model and 54,000 per layout. B
places the rule before the task instruction, M between the task instruction and
the source sentence, and E after the source sentence.
Appendix~\ref{app:reiteration-prompts} specifies the layouts and gives exact
prompts. M and M$\times$5 are the Experiment~1 and five-copy reiteration
responses; the eight additional layouts contribute 432,000 responses, generated
with the same model identifiers and standard settings
(Appendix~\ref{app:inference-details}). All responses are coded with the
Experiment~1 procedure and accepted-form inventory. Seven responses in the
additional layouts (one in E, six in ME) were incomplete and are retained as
\textsc{Failed} in the denominators.

\paragraph{Pooled results.}
Table~\ref{tab:paper-layout-pooled} pools all nine models and both sentence
contexts. Middle repetitions alone do not consistently increase CK: 75.07\%, 75.48\%,
73.23\%, and 71.65\% for one, two, three, and five copies. Placing a copy at
another position does. With two copies, BE gives 86.56\% CK versus 75.48\% for
M$\times$2, an increase of 11.08 percentage points (calculated before rounding)
that occurs in every model (1.60 to 30.88 points) and in every frame; ME gives
84.22\% and BM 78.59\%. With three copies, BME gives 85.24\% versus 73.23\% for
M$\times$3. Under F1, which instructs the model to ignore the rule, BE raises CK
from 61.94\% to 77.83\%; under F5, which requests the supplied meaning, from
86.58\% to 88.63\%.

\paragraph{Baseline-filtered results.}
Filtering follows Appendix~\ref{app:followup-results}: a trial is retained when
the same model produced a \textsc{Familiar} response on the matched no-rule
baseline. Every layout retains 42,270 responses. The filtered rates in
Table~\ref{tab:paper-layout-pooled} preserve every comparison above, and the
direction of each model's change relative to Experiment~1 is unchanged for every
layout. BE gives 84.64\% versus 72.32\% for M$\times$2, with BE higher in every
model; the middle-only layouts give 71.85\%, 72.32\%, 69.86\%, and 68.03\% for
one, two, three, and five copies.

\paragraph{Limits.}
These layout comparisons are descriptive and were not tested statistically.
Middle copies share one input line, whereas copies before the task or after
the source occupy separate lines, so placement and line breaks are confounded.
Lexical inclusion does not establish the fidelity of the rest of the
translation.

\paragraph{Full results by model, frame, and context.}
Table~\ref{tab:paper-layout-full} reports CK rates for every layout and model
in each sentence context. PK and CK denote familiar-compatible and
supplied-compatible sentence contexts. Each frame cell pools 600 responses and
each All cell pools the five frames (3,000 responses); \textsc{Failed}
responses are included in these denominators. Rates count supplied-only
responses; adding \textsc{Both} responses changes no comparison (counts are in
the analysis repository).

% ===== End inline appendix: appendix-position-repetition =====

% Original appendix order: two-column prose, then full-width displays.
% ===== D/E: Full-width figures and statistical tables =====
%
\clearpage
\begin{figure*}[p]
\centering
  \includegraphics[width=\textwidth]{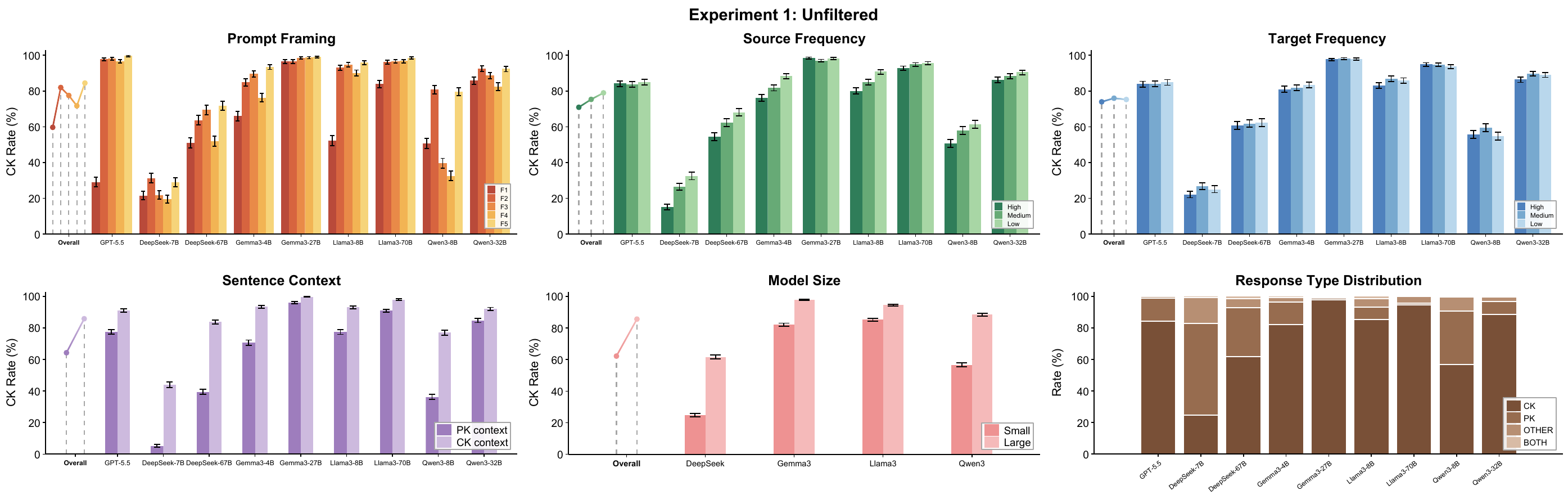}
  \caption{Experiment~1, full dataset ($N=54{,}000$). Target frequency is a proxy (Section~\ref{sec:materials-conditions}).}
  \label{fig:exp1-unfiltered-full}
\par\medskip
\centering
  \includegraphics[width=\textwidth]{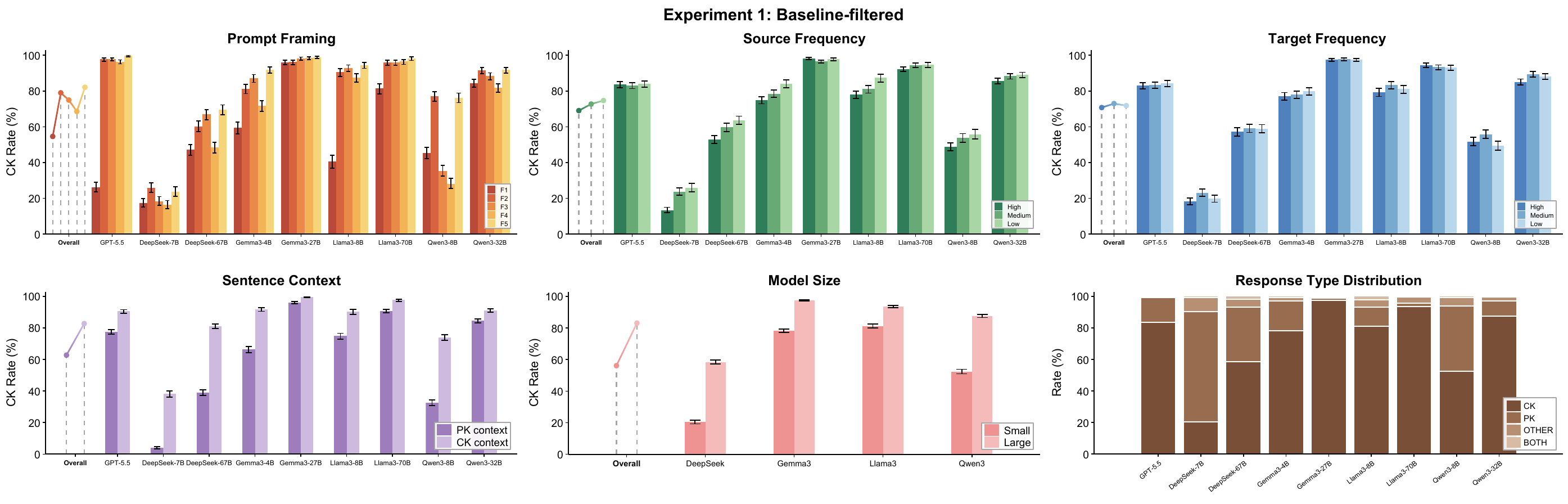}
  \caption{Experiment~1, baseline-filtered dataset ($N=42{,}270$).}
  \label{fig:exp1-filtered}
\par\medskip
\centering
  \includegraphics[width=\textwidth]{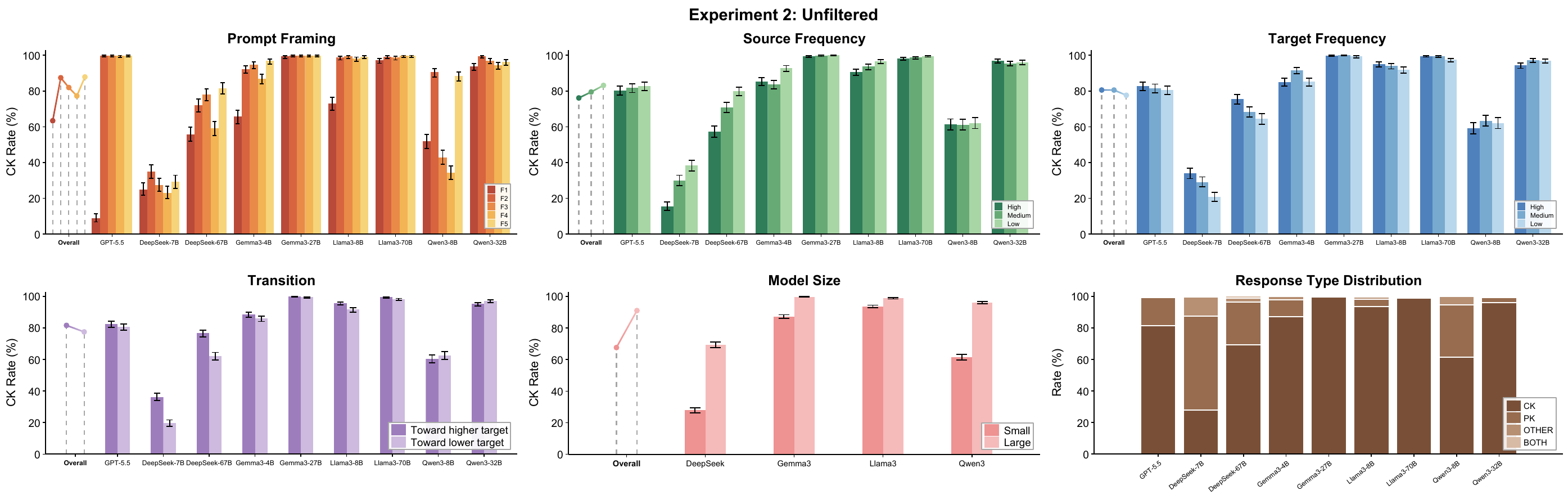}
  \caption{Experiment~2, full dataset ($N=27{,}000$).}
  \label{fig:exp2-unfiltered-full}
\par\medskip
\centering
  \includegraphics[width=\textwidth]{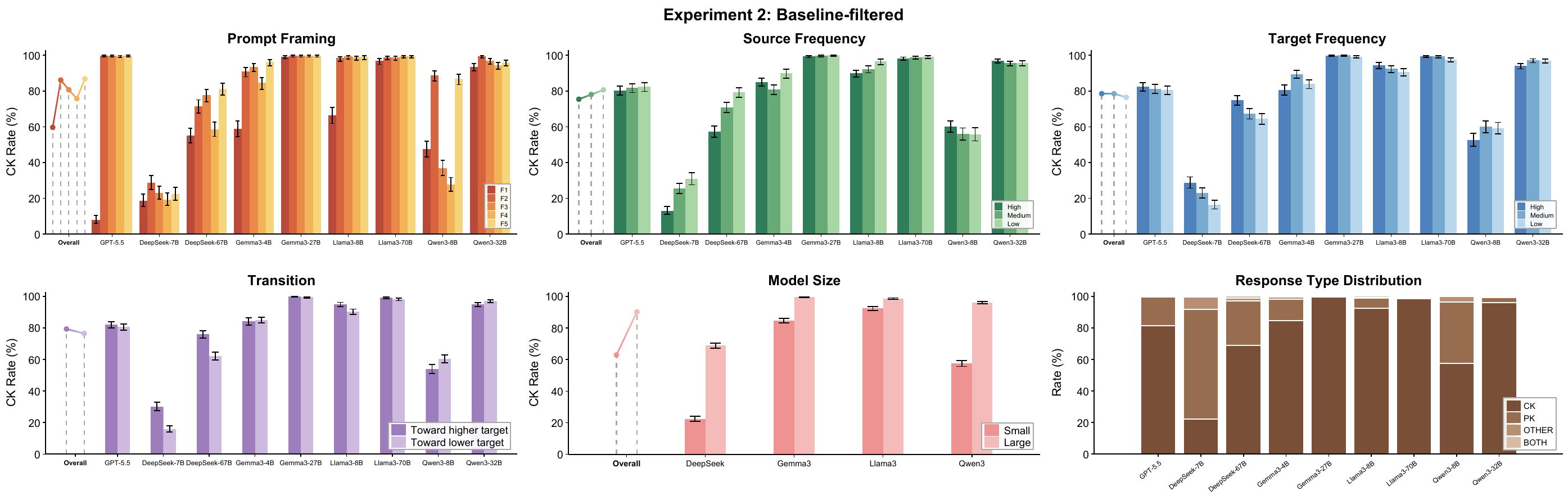}
  \caption{Experiment~2, baseline-filtered dataset ($N=23{,}100$).}
  \label{fig:exp2-filtered}
\end{figure*}

\clearpage
\begin{figure*}[p]
\centering
  \includegraphics[width=\textwidth]{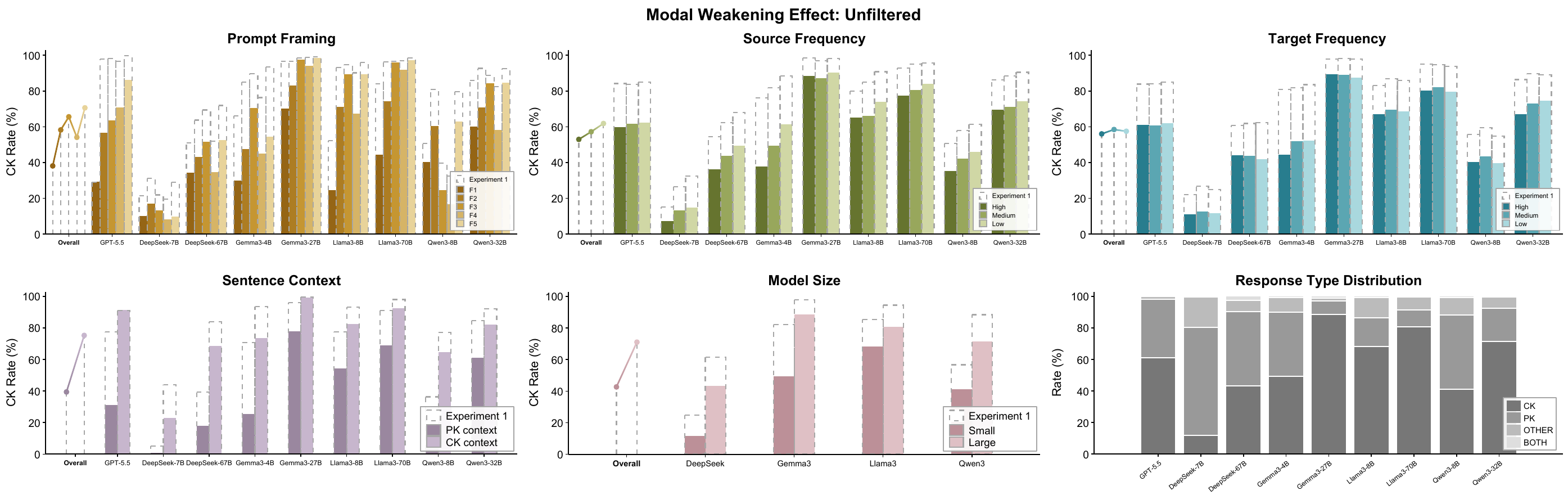}
  \caption{Modal weakening, full dataset ($N=54{,}000$).}
  \label{fig:modal-unfiltered-full}
\par\medskip
\centering
  \includegraphics[width=\textwidth]{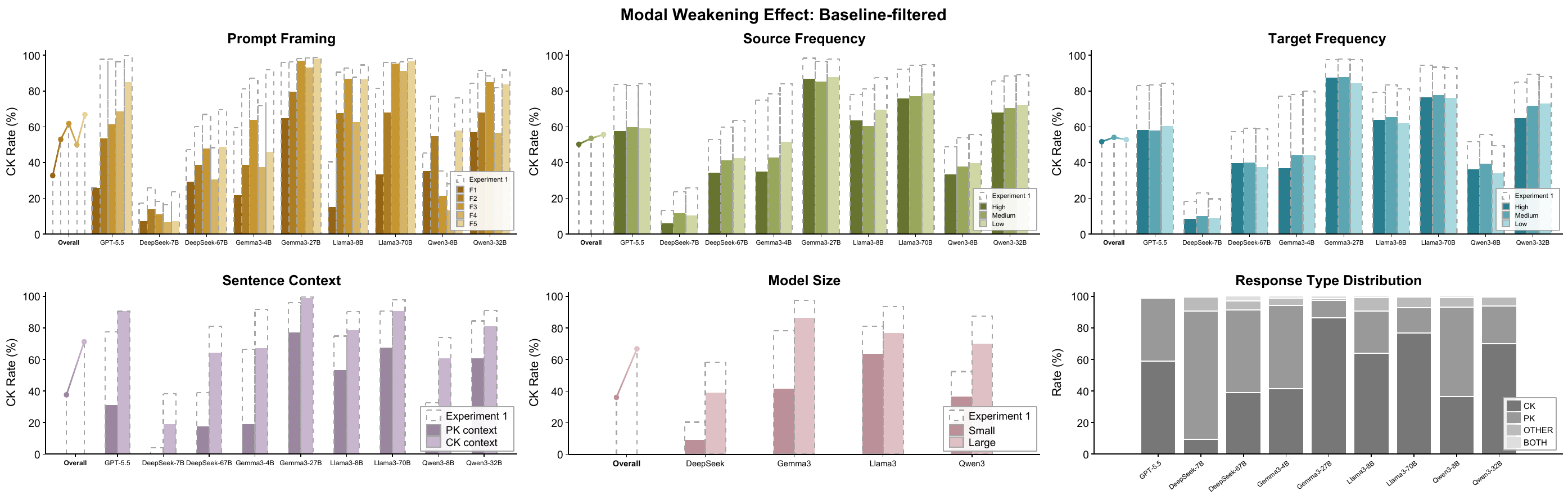}
  \caption{Modal weakening, baseline-filtered dataset ($N=42{,}270$).}
  \label{fig:modal-filtered-full}
\par\medskip
\centering
  \includegraphics[width=\textwidth]{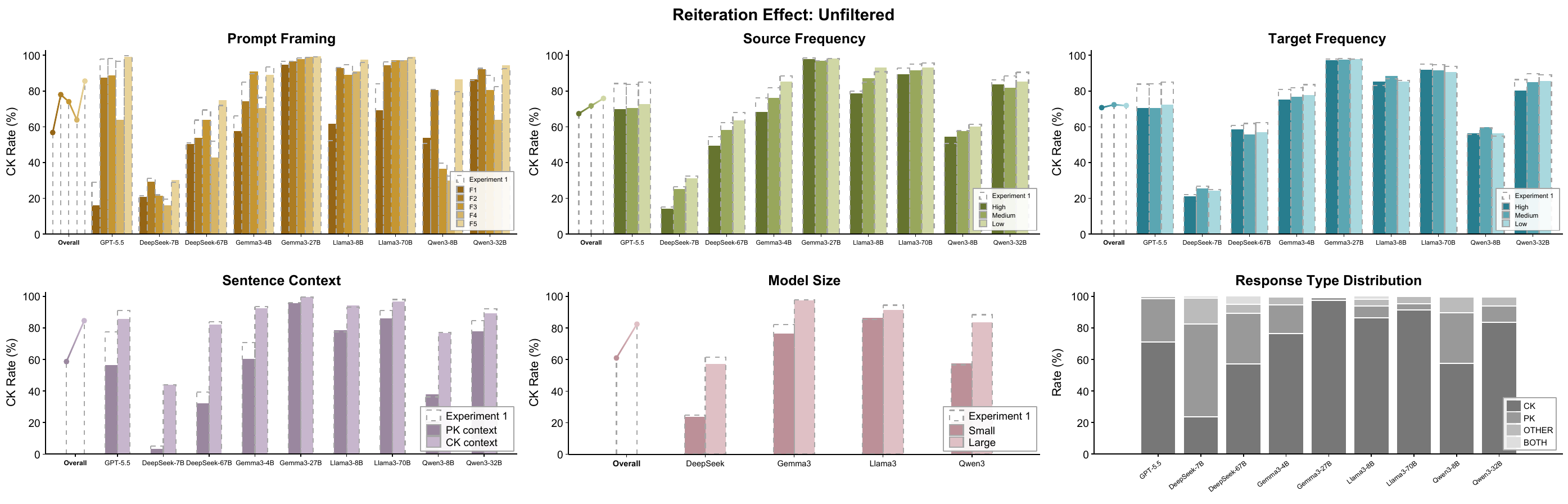}
  \caption{Reiteration (M$\times$5), full dataset ($N=54{,}000$).}
  \label{fig:reiteration-unfiltered-full}
\par\medskip
\centering
  \includegraphics[width=\textwidth]{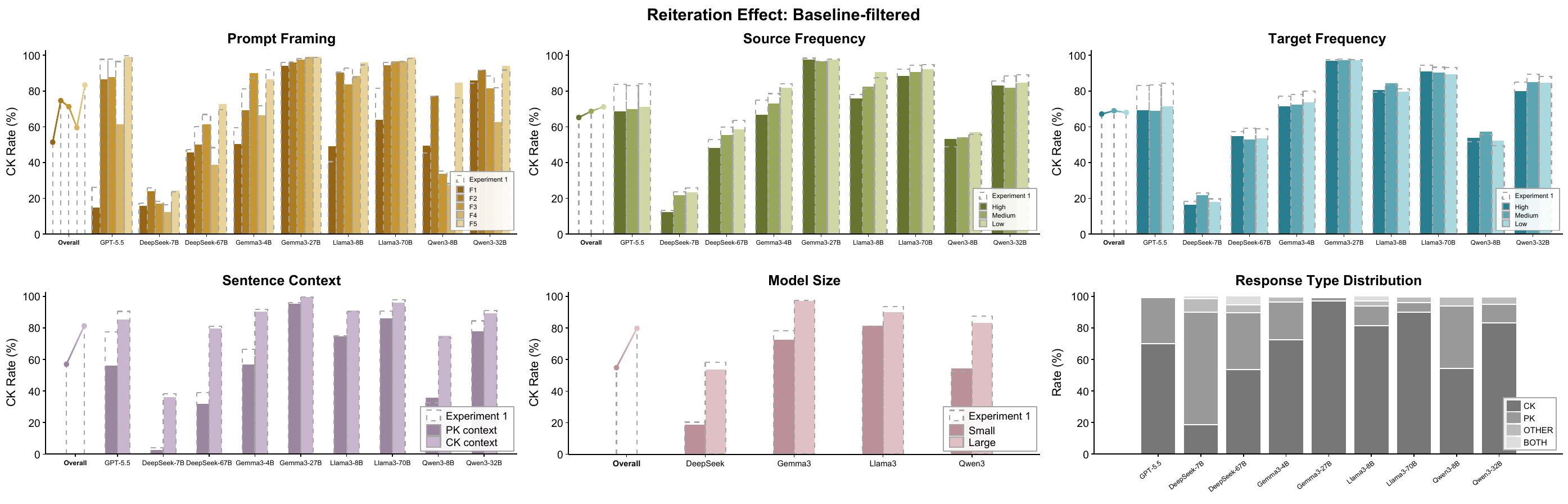}
  \caption{Reiteration (M$\times$5), baseline-filtered dataset ($N=42{,}270$).}
  \label{fig:reiteration-filtered-full}
\end{figure*}

\clearpage
\begin{figure*}[p]
\centering
  \includegraphics[width=\textwidth]{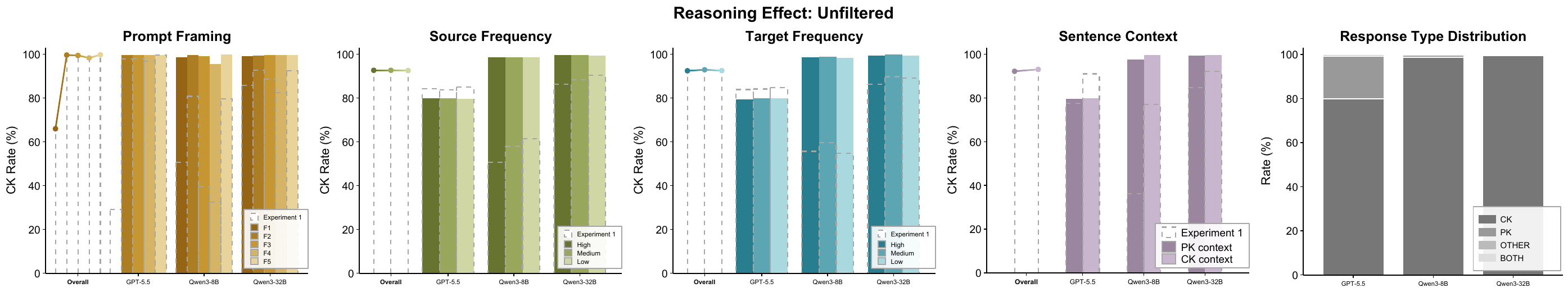}
  \caption{Explicit-reasoning response distributions and CK rates by condition for GPT-5.5, Qwen3-8B, and Qwen3-32B in the full dataset ($N=18{,}000$).}
  \label{fig:reasoning-unfiltered-full}
\par\medskip
\centering
  \includegraphics[width=\textwidth]{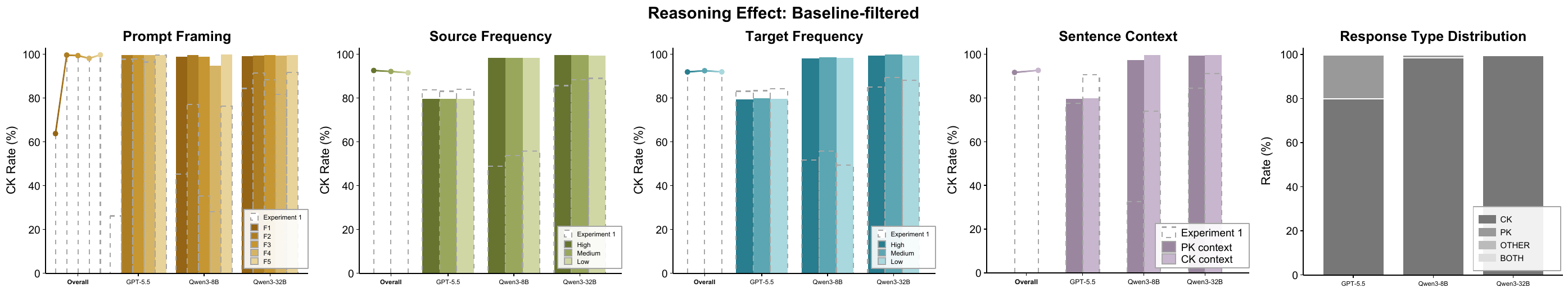}
  \caption{Explicit-reasoning response distributions and CK rates by condition for GPT-5.5, Qwen3-8B, and Qwen3-32B in the baseline-filtered dataset ($N=15{,}440$).}
  \label{fig:reasoning-filtered-full}
\par\medskip
\begin{minipage}{\textwidth}
\centering
\normalsize
\setlength{\tabcolsep}{7pt}
\begin{tabular}{@{}llccc@{}}
  \hline
  \textbf{Experiment} & \textbf{Contrast} & \textbf{OR} & \textbf{95\% CI} & \textbf{$p_{\mathrm{Holm}}$} \\
  \hline
  Exp.~1 & CK context vs. PK context & 7.70 & [7.19, 8.25] & $<.001$ \\
  Exp.~1 & F2 vs. F1 & 7.73 & [7.03, 8.49] & $<.001$ \\
  Exp.~1 & F3 vs. F1 & 5.16 & [4.72, 5.65] & $<.001$ \\
  Exp.~1 & F4 vs. F1 & 2.93 & [2.69, 3.19] & $<.001$ \\
  Exp.~1 & F5 vs. F1 & 10.89 & [9.87, 12.02] & $<.001$ \\
  Exp.~1 & Medium vs. high source frequency & 1.42 & [1.15, 1.76] & $.004$ \\
  Exp.~1 & Low vs. high source frequency & 1.98 & [1.60, 2.46] & $<.001$ \\
  Exp.~1 & Medium vs. high target-frequency proxy & 1.14 & [0.93, 1.40] & $.423$ \\
  Exp.~1 & Low vs. high target-frequency proxy & 1.06 & [0.86, 1.30] & $.584$ \\
  \hline
  Exp.~2 & F2 vs. F1 & 16.07 & [13.81, 18.70] & $<.001$ \\
  Exp.~2 & F3 vs. F1 & 7.76 & [6.76, 8.90] & $<.001$ \\
  Exp.~2 & F4 vs. F1 & 4.44 & [3.90, 5.05] & $<.001$ \\
  Exp.~2 & F5 vs. F1 & 17.57 & [15.07, 20.48] & $<.001$ \\
  Exp.~2 & Medium vs. high source frequency & 1.72 & [1.19, 2.47] & $.011$ \\
  Exp.~2 & Low vs. high source frequency & 2.18 & [1.51, 3.15] & $<.001$ \\
  Exp.~2 & Medium vs. high target-frequency proxy & 1.29 & [0.97, 1.71] & $.169$ \\
  Exp.~2 & Low vs. high target-frequency proxy & 1.06 & [0.80, 1.41] & $.673$ \\
  \hline
\end{tabular}
\captionof{table}{Fixed-effect contrasts from the baseline-filtered GLMMs. Odds ratios predict a \textsc{CK} response. Reference levels are PK context (Experiment~1), F1, and the high bands of source frequency and the target-frequency proxy. $p$ values are Holm-adjusted within each experiment's complete contrast family; 95\% confidence intervals are unadjusted Wald intervals.}
\label{tab:filtered-coefficients}
\end{minipage}
\par\medskip
\begin{minipage}{\textwidth}
\centering
\normalsize
\setlength{\tabcolsep}{6pt}
\begin{tabular}{@{}lcccccc@{}}
  \hline
  & \multicolumn{3}{c}{\textbf{Full-dataset CK rate (\%)}} & \multicolumn{3}{c}{\textbf{Baseline-filtered CK rate (\%)}} \\
  \textbf{Follow-up} & \textbf{Exp.~1} & \textbf{Follow-up} & \textbf{$\Delta$} & \textbf{Exp.~1} & \textbf{Follow-up} & \textbf{$\Delta$} \\
  \hline
  Modal weakening & 75.07 & 57.34 & $-17.73$ & 71.85 & 52.86 & $-18.99$ \\
  Reiteration (M$\times$5) & 75.07 & 71.65 & $-3.42$ & 71.85 & 68.03 & $-3.82$ \\
  Explicit reasoning & 76.48 & 92.65 & $+16.17$ & 75.27 & 92.14 & $+16.88$ \\
  \hline
\end{tabular}
\captionof{table}{Overall CK-rate changes in the controlled follow-ups. $\Delta$ is the follow-up rate minus the matched Experiment~1 rate, in percentage points, calculated before rounding. The reasoning comparison includes only GPT-5.5, Qwen3-8B, and Qwen3-32B. Modal weakening and reiteration each contain 54,000 responses in the full dataset and 42,270 after filtering; reasoning contains 18,000 and 15,440, respectively.}
\label{tab:followup-robustness}
\end{minipage}
\end{figure*}
\clearpage

\onecolumn
% ===== E: Translation Quality tables =====
%
\begin{table}[p]
\centering
{\normalsize
\setlength{\tabcolsep}{3pt}
\begin{tabular}{@{}>{\raggedright\arraybackslash}p{5.0cm}*{10}{r}@{}}
\hline
 & \multicolumn{2}{c}{\textbf{F1}} & \multicolumn{2}{c}{\textbf{F2}} & \multicolumn{2}{c}{\textbf{F3}} & \multicolumn{2}{c}{\textbf{F4}} & \multicolumn{2}{c}{\textbf{F5}} \\
\cline{2-3}\cline{4-5}\cline{6-7}\cline{8-9}\cline{10-11}
\textbf{Condition} & \textbf{Fam.} & \textbf{Sup.} & \textbf{Fam.} & \textbf{Sup.} & \textbf{Fam.} & \textbf{Sup.} & \textbf{Fam.} & \textbf{Sup.} & \textbf{Fam.} & \textbf{Sup.} \\
\hline
\multicolumn{11}{@{}l}{\textbf{Baselines (Familiar reference only)}}\\*
No rule: semantically consistent & \multicolumn{10}{c}{0.909 (no frame)} \\
No rule: semantically inconsistent & \multicolumn{10}{c}{0.813 (no frame)} \\
Conventional-meaning rule & 0.924 & -- & 0.924 & -- & 0.911 & -- & 0.910 & -- & 0.924 & -- \\
\hline
\multicolumn{11}{@{}l}{\textbf{Main experiments and modal weakening}}\\*
Experiment 1 & 0.767 & \textbf{0.826} & 0.723 & \textbf{0.862} & 0.721 & \textbf{0.842} & 0.732 & \textbf{0.835} & 0.714 & \textbf{0.862} \\
Experiment 2 & 0.852 & \textbf{0.888} & 0.830 & \textbf{0.910} & 0.824 & \textbf{0.893} & 0.828 & \textbf{0.890} & 0.828 & \textbf{0.909} \\
Modal weakening & \textbf{0.812} & 0.794 & 0.776 & \textbf{0.830} & 0.747 & \textbf{0.827} & 0.765 & \textbf{0.807} & 0.742 & \textbf{0.840} \\
\hline
\multicolumn{11}{@{}l}{\textbf{Position and repetition}}\\*
Before task (B) & 0.773 & \textbf{0.826} & 0.742 & \textbf{0.848} & 0.716 & \textbf{0.854} & 0.729 & \textbf{0.847} & 0.723 & \textbf{0.853} \\
Middle (M; Experiment 1) & 0.767 & \textbf{0.826} & 0.723 & \textbf{0.862} & 0.721 & \textbf{0.842} & 0.732 & \textbf{0.835} & 0.714 & \textbf{0.862} \\
After source (E) & 0.772 & \textbf{0.823} & 0.740 & \textbf{0.847} & 0.702 & \textbf{0.851} & 0.712 & \textbf{0.838} & 0.718 & \textbf{0.851} \\
Two middle copies (M$\times$2) & 0.764 & \textbf{0.830} & 0.724 & \textbf{0.863} & 0.721 & \textbf{0.840} & 0.727 & \textbf{0.826} & 0.707 & \textbf{0.864} \\
Before task + middle (BM) & 0.748 & \textbf{0.825} & 0.723 & \textbf{0.842} & 0.699 & \textbf{0.849} & 0.712 & \textbf{0.852} & 0.707 & \textbf{0.861} \\
Before task + after source (BE) & 0.720 & \textbf{0.846} & 0.707 & \textbf{0.863} & 0.689 & \textbf{0.861} & 0.692 & \textbf{0.862} & 0.697 & \textbf{0.864} \\
Middle + after source (ME) & 0.729 & \textbf{0.843} & 0.705 & \textbf{0.865} & 0.698 & \textbf{0.850} & 0.695 & \textbf{0.838} & 0.697 & \textbf{0.868} \\
Three middle copies (M$\times$3) & 0.768 & \textbf{0.826} & 0.727 & \textbf{0.860} & 0.722 & \textbf{0.835} & 0.735 & \textbf{0.821} & 0.708 & \textbf{0.863} \\
Before task + middle + after source (BME) & 0.716 & \textbf{0.829} & 0.702 & \textbf{0.848} & 0.679 & \textbf{0.847} & 0.692 & \textbf{0.851} & 0.695 & \textbf{0.862} \\
Five middle copies (M$\times$5) & 0.772 & \textbf{0.819} & 0.732 & \textbf{0.857} & 0.729 & \textbf{0.840} & 0.738 & \textbf{0.815} & 0.710 & \textbf{0.863} \\
\hline
\multicolumn{11}{@{}l}{\textbf{Explicit reasoning: GPT-5.5, Qwen3-8B and Qwen3-32B}}\\*
Experiment 1, same models & 0.789 & \textbf{0.825} & 0.713 & \textbf{0.886} & 0.735 & \textbf{0.857} & 0.745 & \textbf{0.849} & 0.706 & \textbf{0.880} \\
Explicit reasoning & 0.750 & \textbf{0.823} & 0.681 & \textbf{0.888} & 0.680 & \textbf{0.885} & 0.682 & \textbf{0.885} & 0.680 & \textbf{0.887} \\
\hline
\end{tabular}
\caption{COMET mean sentence scores (0--1 scale) by condition and frame, pooling sentence contexts and all nine models unless a block names its models. Within each frame, \textbf{bold} marks the reference against which the outputs score higher; baseline rows have one reference.}\label{tab:paper-metric-comet}
}
\end{table}

\clearpage
\begin{table}[p]
\centering
{\normalsize
\setlength{\tabcolsep}{3pt}
\begin{tabular}{@{}>{\raggedright\arraybackslash}p{5.0cm}*{10}{r}@{}}
\hline
 & \multicolumn{2}{c}{\textbf{F1}} & \multicolumn{2}{c}{\textbf{F2}} & \multicolumn{2}{c}{\textbf{F3}} & \multicolumn{2}{c}{\textbf{F4}} & \multicolumn{2}{c}{\textbf{F5}} \\
\cline{2-3}\cline{4-5}\cline{6-7}\cline{8-9}\cline{10-11}
\textbf{Condition} & \textbf{Fam.} & \textbf{Sup.} & \textbf{Fam.} & \textbf{Sup.} & \textbf{Fam.} & \textbf{Sup.} & \textbf{Fam.} & \textbf{Sup.} & \textbf{Fam.} & \textbf{Sup.} \\
\hline
\multicolumn{11}{@{}l}{\textbf{Baselines (Familiar reference only)}}\\*
No rule: semantically consistent & \multicolumn{10}{c}{71.26 (no frame)} \\
No rule: semantically inconsistent & \multicolumn{10}{c}{63.35 (no frame)} \\
Conventional-meaning rule & 75.44 & -- & 74.44 & -- & 72.36 & -- & 71.64 & -- & 74.47 & -- \\
\hline
\multicolumn{11}{@{}l}{\textbf{Main experiments and modal weakening}}\\*
Experiment 1 & 61.82 & \textbf{66.99} & 58.64 & \textbf{69.97} & 57.18 & \textbf{67.37} & 57.88 & \textbf{66.60} & 58.19 & \textbf{70.15} \\
Experiment 2 & 63.80 & \textbf{69.85} & 59.48 & \textbf{72.64} & 58.79 & \textbf{70.38} & 58.82 & \textbf{69.11} & 59.67 & \textbf{73.01} \\
Modal weakening & \textbf{64.45} & 63.93 & 61.46 & \textbf{66.26} & 58.82 & \textbf{65.70} & 59.64 & \textbf{63.90} & 59.55 & \textbf{67.88} \\
\hline
\multicolumn{11}{@{}l}{\textbf{Position and repetition}}\\*
Before task (B) & 62.07 & \textbf{66.65} & 59.65 & \textbf{68.40} & 56.75 & \textbf{68.09} & 57.98 & \textbf{67.72} & 58.60 & \textbf{69.13} \\
Middle (M; Experiment 1) & 61.82 & \textbf{66.99} & 58.64 & \textbf{69.97} & 57.18 & \textbf{67.37} & 57.88 & \textbf{66.60} & 58.19 & \textbf{70.15} \\
After source (E) & 61.73 & \textbf{66.46} & 59.76 & \textbf{68.76} & 55.62 & \textbf{67.77} & 56.14 & \textbf{66.20} & 58.28 & \textbf{69.28} \\
Two middle copies (M$\times$2) & 61.18 & \textbf{66.90} & 58.69 & \textbf{69.95} & 56.88 & \textbf{66.71} & 57.11 & \textbf{65.52} & 57.53 & \textbf{69.96} \\
Before task + middle (BM) & 59.24 & \textbf{65.63} & 57.29 & \textbf{66.90} & 55.43 & \textbf{67.35} & 56.85 & \textbf{68.10} & 57.08 & \textbf{69.23} \\
Before task + after source (BE) & 57.83 & \textbf{67.76} & 57.41 & \textbf{69.87} & 55.22 & \textbf{68.56} & 55.93 & \textbf{69.29} & 56.25 & \textbf{69.24} \\
Middle + after source (ME) & 59.33 & \textbf{68.50} & 57.32 & \textbf{70.14} & 55.34 & \textbf{67.68} & 55.14 & \textbf{66.61} & 57.00 & \textbf{70.48} \\
Three middle copies (M$\times$3) & 61.35 & \textbf{66.43} & 58.99 & \textbf{69.70} & 57.27 & \textbf{66.60} & 57.53 & \textbf{64.86} & 57.52 & \textbf{69.80} \\
Before task + middle + after source (BME) & 56.14 & \textbf{65.15} & 55.67 & \textbf{67.31} & 53.56 & \textbf{66.69} & 55.17 & \textbf{67.58} & 55.86 & \textbf{68.93} \\
Five middle copies (M$\times$5) & 61.31 & \textbf{65.69} & 59.28 & \textbf{69.43} & 58.63 & \textbf{67.87} & 58.59 & \textbf{65.22} & 57.85 & \textbf{70.04} \\
\hline
\multicolumn{11}{@{}l}{\textbf{Explicit reasoning: GPT-5.5, Qwen3-8B and Qwen3-32B}}\\*
Experiment 1, same models & 64.37 & \textbf{67.56} & 60.31 & \textbf{74.55} & 60.79 & \textbf{71.29} & 61.71 & \textbf{71.00} & 59.00 & \textbf{73.24} \\
Explicit reasoning & 61.34 & \textbf{66.87} & 56.57 & \textbf{73.17} & 56.08 & \textbf{72.43} & 56.80 & \textbf{72.70} & 55.83 & \textbf{72.31} \\
\hline
\end{tabular}
\caption{chrF++ mean sentence scores (0--100 scale) by condition and frame, pooling sentence contexts and all nine models unless a block names its models. Within each frame, \textbf{bold} marks the reference against which the outputs score higher; baseline rows have one reference.}\label{tab:paper-metric-chrf}
}
\end{table}
\twocolumn

\raggedbottom
\onecolumn
% ===== F: Conventional-meaning baseline table =====
%
\begin{center}
\begin{minipage}{\textwidth}
\centering
\normalsize
\setlength{\tabcolsep}{6pt}
\begin{tabular}{@{}lrrrrrrr@{}}
\hline
 & & \multicolumn{6}{c@{}}{\textbf{Agreeing rule}} \\
\textbf{Model} & \textbf{No rule} & \textbf{F1} & \textbf{F2} & \textbf{F3} & \textbf{F4} & \textbf{F5} & \textbf{All} \\
\hline
GPT-5.5 & 98.00 & 100.00 & 100.00 & 100.00 & 100.00 & 100.00 & 100.00 \\
DeepSeek-LLM-7B & 84.00 & 96.67 & 97.33 & 93.33 & 93.33 & 96.67 & 95.47 \\
DeepSeek-LLM-67B & 96.00 & 98.67 & 99.33 & 84.00 & 90.00 & 99.33 & 94.27 \\
Gemma-3-4B & 78.67 & 100.00 & 100.00 & 100.00 & 100.00 & 100.00 & 100.00 \\
Gemma-3-27B & 90.00 & 100.00 & 100.00 & 100.00 & 100.00 & 100.00 & 100.00 \\
Llama-3-8B & 72.00 & 100.00 & 100.00 & 100.00 & 100.00 & 100.00 & 100.00 \\
Llama-3-70B & 75.33 & 99.33 & 100.00 & 99.33 & 100.00 & 100.00 & 99.73 \\
Qwen3-8B & 83.33 & 98.67 & 100.00 & 98.67 & 99.33 & 100.00 & 99.33 \\
Qwen3-32B & 92.67 & 100.00 & 100.00 & 100.00 & 100.00 & 100.00 & 100.00 \\
\textbf{Pooled} & 85.56 & 99.26 & 99.63 & 97.26 & 98.07 & 99.56 & 98.76 \\
\hline
\end{tabular}
\captionof{table}{Familiar-form inclusion (\%) without a lexical rule and with a rule agreeing with the familiar meaning. No rule: 150 semantically consistent responses per model. Each frame: 150 attempts per model; All pools the 750. Incomplete responses count as misses; completed-only rates are given in the text.}\label{tab:paper-control-full}
\end{minipage}
\end{center}
\par\medskip
% ===== G: Reiteration and position tables =====
%
{\normalsize
\setlength{\LTcapwidth}{\textwidth}
\setlength{\tabcolsep}{6pt}
\begin{longtable}{@{}lrrrrrrr@{}}
\hline
\textbf{Layout} & \textbf{CK} & \textbf{Filtered} & \textbf{F1} & \textbf{F2} & \textbf{F3} & \textbf{F4} & \textbf{F5} \\
\hline
\endfirsthead
\multicolumn{8}{l}{\tablename\ \thetable\ continued}\\
\hline
\textbf{Layout} & \textbf{CK} & \textbf{Filtered} & \textbf{F1} & \textbf{F2} & \textbf{F3} & \textbf{F4} & \textbf{F5} \\
\hline
\endhead
\hline
\endfoot
\endlastfoot
B & 73.55 & 70.14 & 58.36 & 72.63 & \textbf{81.59} & 75.89 & 79.27 \\
M & 75.07 & 71.85 & 59.70 & 81.96 & 77.52 & 71.72 & \textbf{84.46} \\
E & 75.18 & 71.88 & 58.18 & 73.26 & \textbf{85.64} & 78.16 & 80.65 \\
M$\times$2 & 75.48 & 72.32 & 61.94 & 81.70 & 76.56 & 70.62 & \textbf{86.58} \\
BM & 78.59 & 75.74 & 65.11 & 76.29 & 84.65 & 81.41 & \textbf{85.48} \\
BE & 86.56 & 84.64 & 77.83 & 85.97 & 90.10 & \textbf{90.27} & 88.63 \\
ME & 84.22 & 82.03 & 75.13 & 86.98 & 85.92 & 82.77 & \textbf{90.30} \\
M$\times$3 & 73.23 & 69.86 & 59.56 & 80.06 & 74.00 & 66.59 & \textbf{85.93} \\
BME & 85.24 & 83.11 & 75.67 & 83.65 & \textbf{90.64} & 86.75 & 89.52 \\
M$\times$5 & 71.65 & 68.03 & 56.81 & 77.99 & 74.04 & 63.85 & \textbf{85.56} \\
\hline
\caption{CK rates (\%) for the ten layouts, pooled across nine models. CK and F1--F5 use the full dataset (54,000 responses per layout; 10,800 per frame). Filtered uses the 42,270 baseline-filtered responses per layout. CK counts supplied-only responses. In each row, \textbf{bold} marks the highest frame rate.}\label{tab:paper-layout-pooled}\\
\end{longtable}
}

\par\medskip
{\normalsize
\setlength{\LTcapwidth}{\textwidth}
\setlength{\tabcolsep}{3pt}
\begin{longtable}{@{}l rrrrrr rrrrrr@{}}
\hline
& \multicolumn{6}{c}{\textbf{PK context}} & \multicolumn{6}{c@{}}{\textbf{CK context}} \\
\textbf{Layout} & \textbf{All} & \textbf{F1} & \textbf{F2} & \textbf{F3} & \textbf{F4} & \textbf{F5} & \textbf{All} & \textbf{F1} & \textbf{F2} & \textbf{F3} & \textbf{F4} & \textbf{F5} \\
\hline
\endfirsthead
\multicolumn{13}{l}{\tablename\ \thetable\ continued}\\
\hline
& \multicolumn{6}{c}{\textbf{PK context}} & \multicolumn{6}{c@{}}{\textbf{CK context}} \\
\textbf{Layout} & \textbf{All} & \textbf{F1} & \textbf{F2} & \textbf{F3} & \textbf{F4} & \textbf{F5} & \textbf{All} & \textbf{F1} & \textbf{F2} & \textbf{F3} & \textbf{F4} & \textbf{F5} \\
\hline
\endhead
\hline
\endfoot
\endlastfoot
\multicolumn{13}{@{}l}{\textbf{GPT-5.5}}\\*
B & 74.93 & 0.33 & 96.67 & 91.17 & 88.00 & \textbf{98.50} & 89.07 & 46.17 & 99.67 & \textbf{99.83} & \textbf{99.83} & \textbf{99.83} \\
M & 77.60 & 2.33 & 96.00 & 96.33 & 93.67 & \textbf{99.67} & 91.03 & 55.83 & \textbf{99.83} & \textbf{99.83} & \textbf{99.83} & \textbf{99.83} \\
E & 69.40 & 4.00 & 77.67 & 94.50 & 71.33 & \textbf{99.50} & 94.87 & 75.00 & \textbf{99.83} & \textbf{99.83} & \textbf{99.83} & \textbf{99.83} \\
M$\times$2 & 76.33 & 0.67 & 93.33 & 96.17 & 91.83 & \textbf{99.67} & 91.53 & 58.33 & \textbf{99.83} & \textbf{99.83} & \textbf{99.83} & \textbf{99.83} \\
BM & 76.97 & 0.67 & 90.17 & 96.67 & 98.17 & \textbf{99.17} & 89.90 & 50.17 & \textbf{99.83} & \textbf{99.83} & \textbf{99.83} & \textbf{99.83} \\
BE & 80.03 & 2.67 & 99.00 & \textbf{99.83} & 98.83 & \textbf{99.83} & 91.03 & 55.83 & \textbf{99.83} & \textbf{99.83} & \textbf{99.83} & \textbf{99.83} \\
ME & 79.70 & 0.67 & 99.00 & 99.50 & \textbf{99.83} & 99.50 & 92.63 & 63.83 & \textbf{99.83} & \textbf{99.83} & \textbf{99.83} & \textbf{99.83} \\
M$\times$3 & 69.43 & 0.50 & 89.00 & 92.17 & 66.33 & \textbf{99.17} & 90.27 & 52.17 & \textbf{99.83} & \textbf{99.83} & 99.67 & \textbf{99.83} \\
BME & 79.03 & 1.83 & 94.33 & 99.67 & 99.50 & \textbf{99.83} & 88.20 & 41.67 & \textbf{99.83} & \textbf{99.83} & \textbf{99.83} & \textbf{99.83} \\
M$\times$5 & 56.33 & 0.00 & 75.00 & 77.50 & 30.83 & \textbf{98.33} & 85.80 & 32.50 & \textbf{99.83} & \textbf{99.83} & 97.00 & \textbf{99.83} \\
\hline
\multicolumn{13}{@{}l}{\textbf{DeepSeek-LLM-7B}}\\*
B & 2.90 & \textbf{5.50} & 2.33 & 1.83 & 1.17 & 3.67 & 42.27 & 47.17 & 37.83 & 41.17 & 33.50 & \textbf{51.67} \\
M & 5.20 & 4.33 & 6.33 & 4.83 & 2.83 & \textbf{7.67} & 44.10 & 38.67 & \textbf{56.33} & 39.00 & 36.17 & 50.33 \\
E & 13.87 & 3.83 & 5.67 & \textbf{29.17} & 25.83 & 4.83 & 44.97 & 26.17 & 43.50 & 58.33 & \textbf{67.50} & 29.33 \\
M$\times$2 & 7.17 & 6.17 & 10.83 & 4.00 & 2.00 & \textbf{12.83} & 53.67 & 49.83 & \textbf{67.50} & 50.67 & 39.17 & 61.17 \\
BM & 8.33 & \textbf{14.67} & 9.17 & 9.33 & 1.83 & 6.67 & 47.83 & 55.50 & 45.50 & 41.83 & 38.17 & \textbf{58.17} \\
BE & 9.70 & \textbf{12.83} & 6.17 & 9.33 & 11.50 & 8.67 & 55.10 & 54.00 & 46.00 & 58.17 & 52.17 & \textbf{65.17} \\
ME & 9.20 & 8.83 & 9.00 & 6.17 & 7.50 & \textbf{14.50} & 56.30 & 53.67 & 63.50 & 54.33 & 46.00 & \textbf{64.00} \\
M$\times$3 & 4.83 & 5.67 & 6.67 & 1.50 & 0.67 & \textbf{9.67} & 47.33 & 44.50 & \textbf{62.33} & 40.50 & 31.83 & 57.50 \\
BME & 17.20 & \textbf{25.33} & 16.00 & 22.50 & 6.83 & 15.33 & 58.07 & 62.00 & 52.00 & 60.83 & 50.17 & \textbf{65.33} \\
M$\times$5 & 3.17 & 2.17 & 3.83 & 1.67 & 0.50 & \textbf{7.67} & 44.10 & 39.67 & \textbf{55.00} & 41.33 & 31.83 & 52.67 \\
\hline
\multicolumn{13}{@{}l}{\textbf{DeepSeek-LLM-67B}}\\*
B & 38.70 & 21.83 & 23.33 & 49.67 & \textbf{58.33} & 40.33 & 87.10 & 81.17 & 85.33 & \textbf{92.83} & 90.67 & 85.50 \\
M & 39.50 & 26.00 & 37.83 & 48.67 & 29.00 & \textbf{56.00} & 83.80 & 76.17 & 89.67 & \textbf{90.17} & 75.17 & 87.83 \\
E & 41.27 & 14.00 & 22.50 & \textbf{61.17} & 56.33 & 52.33 & 85.97 & 77.50 & \textbf{90.17} & 89.50 & 89.00 & 83.67 \\
M$\times$2 & 35.27 & 26.00 & 30.83 & 38.33 & 21.83 & \textbf{59.33} & 84.40 & 82.83 & 88.83 & 89.00 & 69.00 & \textbf{92.33} \\
BM & 37.53 & 18.67 & 17.33 & 49.17 & \textbf{53.00} & 49.50 & 86.57 & 81.00 & 80.33 & 89.83 & \textbf{91.17} & 90.50 \\
BE & 84.57 & 74.50 & 74.33 & 91.50 & \textbf{94.83} & 87.67 & 96.87 & 95.00 & 95.17 & 98.00 & \textbf{99.17} & 97.00 \\
ME & 69.60 & 59.83 & 56.17 & 75.83 & 73.00 & \textbf{83.17} & 93.53 & 96.00 & 95.33 & 91.33 & 87.17 & \textbf{97.83} \\
M$\times$3 & 30.97 & 21.17 & 23.33 & 30.50 & 21.83 & \textbf{58.00} & 81.33 & 79.83 & 85.67 & 79.67 & 69.50 & \textbf{92.00} \\
BME & 62.27 & 40.33 & 33.50 & \textbf{81.00} & 79.33 & 77.17 & 90.97 & 88.50 & 88.67 & 93.17 & 89.83 & \textbf{94.67} \\
M$\times$5 & 32.03 & 20.00 & 21.50 & 39.50 & 21.67 & \textbf{57.50} & 82.27 & 80.50 & 86.33 & 88.50 & 63.83 & \textbf{92.17} \\
\hline
\multicolumn{13}{@{}l}{\textbf{Gemma-3-4B}}\\*
B & 61.27 & 36.83 & 56.50 & 75.50 & 54.17 & \textbf{83.33} & 92.33 & 79.17 & 93.83 & \textbf{97.83} & 94.83 & 96.00 \\
M & 70.73 & 47.50 & 75.50 & 81.67 & 59.50 & \textbf{89.50} & 93.57 & 84.67 & 94.50 & \textbf{97.83} & 93.17 & 97.67 \\
E & 42.70 & 8.00 & 22.67 & \textbf{82.33} & 35.83 & 64.67 & 78.03 & 57.33 & 80.50 & \textbf{93.00} & 79.33 & 80.00 \\
M$\times$2 & 64.17 & 38.33 & 65.50 & 80.00 & 53.17 & \textbf{83.83} & 94.23 & 86.33 & 93.83 & \textbf{98.50} & 95.17 & 97.33 \\
BM & 82.80 & 61.00 & 82.67 & \textbf{94.67} & 85.00 & 90.67 & 96.20 & 90.33 & 95.83 & \textbf{98.83} & 98.33 & 97.67 \\
BE & 94.40 & 88.33 & 92.50 & \textbf{99.00} & 96.67 & 95.50 & 97.70 & 95.17 & 97.17 & \textbf{99.50} & 99.33 & 97.33 \\
ME & 90.50 & 75.33 & 92.83 & \textbf{98.83} & 88.00 & 97.50 & 97.03 & 93.83 & 96.50 & \textbf{99.17} & 97.33 & 98.33 \\
M$\times$3 & 61.00 & 33.50 & 60.00 & 81.50 & 47.83 & \textbf{82.17} & 93.63 & 85.17 & 93.67 & \textbf{97.83} & 94.00 & 97.50 \\
BME & 95.10 & 90.33 & 93.83 & \textbf{99.00} & 96.17 & 96.17 & 98.10 & 95.83 & 97.83 & \textbf{99.50} & 99.00 & 98.33 \\
M$\times$5 & 60.57 & 32.83 & 57.00 & \textbf{84.33} & 47.83 & 80.83 & 92.50 & 82.50 & 91.67 & \textbf{97.50} & 93.50 & 97.33 \\
\hline
\multicolumn{13}{@{}l}{\textbf{Gemma-3-27B}}\\*
B & 96.43 & 94.00 & 96.17 & 96.67 & 97.50 & \textbf{97.83} & 99.50 & 99.00 & \textbf{99.67} & \textbf{99.67} & 99.50 & \textbf{99.67} \\
M & 96.17 & 93.50 & 93.50 & 97.50 & 97.83 & \textbf{98.50} & 99.77 & 99.67 & \textbf{99.83} & 99.67 & \textbf{99.83} & \textbf{99.83} \\
E & 96.93 & 95.17 & 95.67 & 98.50 & 96.33 & \textbf{99.00} & 99.57 & 99.33 & 99.50 & 99.67 & \textbf{99.83} & 99.50 \\
M$\times$2 & 95.50 & 91.67 & 92.33 & 96.83 & \textbf{98.33} & \textbf{98.33} & 99.70 & 99.50 & \textbf{99.83} & 99.50 & \textbf{99.83} & \textbf{99.83} \\
BM & 97.90 & 97.00 & 96.00 & 98.17 & \textbf{99.33} & 99.00 & 99.50 & 99.33 & 99.50 & 99.50 & 99.50 & \textbf{99.67} \\
BE & 99.50 & 99.67 & 99.33 & \textbf{99.83} & 99.50 & 99.17 & 99.70 & 99.67 & \textbf{99.83} & \textbf{99.83} & 99.50 & 99.67 \\
ME & 98.77 & 98.67 & 98.50 & 98.50 & 98.83 & \textbf{99.33} & 99.80 & 99.67 & \textbf{99.83} & \textbf{99.83} & \textbf{99.83} & \textbf{99.83} \\
M$\times$3 & 95.67 & 91.33 & 94.17 & 96.00 & 97.83 & \textbf{99.00} & 99.67 & 99.33 & \textbf{99.83} & 99.50 & \textbf{99.83} & \textbf{99.83} \\
BME & 99.03 & 99.00 & 98.67 & 99.00 & \textbf{99.67} & 98.83 & 99.63 & 99.50 & \textbf{99.67} & \textbf{99.67} & \textbf{99.67} & \textbf{99.67} \\
M$\times$5 & 95.57 & 90.33 & 93.50 & 96.33 & 98.67 & \textbf{99.00} & 99.67 & 99.33 & \textbf{99.83} & 99.50 & \textbf{99.83} & \textbf{99.83} \\
\hline
\multicolumn{13}{@{}l}{\textbf{Llama-3-8B}}\\*
B & 75.93 & 45.00 & 71.00 & \textbf{92.83} & 79.50 & 91.33 & 94.30 & 85.00 & 94.00 & \textbf{99.33} & 96.50 & 96.67 \\
M & 77.43 & 31.00 & 88.33 & 90.67 & 83.17 & \textbf{94.00} & 93.20 & 73.67 & 98.17 & \textbf{99.00} & 97.17 & 98.00 \\
E & 75.33 & 38.17 & 73.33 & \textbf{92.33} & 84.50 & 88.33 & 88.87 & 85.83 & \textbf{93.67} & 93.50 & 78.00 & 93.33 \\
M$\times$2 & 82.57 & 47.00 & 88.17 & 90.17 & 91.00 & \textbf{96.50} & 95.53 & 84.50 & 98.17 & \textbf{99.50} & 96.83 & 98.67 \\
BM & 84.87 & 48.83 & 88.17 & \textbf{96.50} & 95.50 & 95.33 & 95.70 & 85.50 & 97.00 & \textbf{99.33} & 98.67 & 98.00 \\
BE & 93.67 & 90.33 & 91.33 & \textbf{96.00} & 95.50 & 95.17 & 98.50 & 97.00 & 99.00 & \textbf{99.33} & 99.17 & 98.00 \\
ME & 86.87 & 72.67 & 80.33 & \textbf{96.17} & 89.83 & 95.33 & 95.47 & 90.50 & 96.50 & 97.83 & 93.83 & \textbf{98.67} \\
M$\times$3 & 81.17 & 45.67 & 88.83 & 85.17 & 89.67 & \textbf{96.50} & 95.10 & 84.17 & 98.50 & \textbf{99.17} & 94.83 & 98.83 \\
BME & 91.70 & 77.67 & 91.00 & \textbf{97.17} & 96.83 & 95.83 & 97.77 & 94.83 & 98.33 & \textbf{98.83} & 98.50 & 98.33 \\
M$\times$5 & 78.57 & 41.17 & 88.17 & 80.67 & 86.33 & \textbf{96.50} & 94.20 & 82.33 & 97.83 & 97.17 & 95.33 & \textbf{98.33} \\
\hline
\multicolumn{13}{@{}l}{\textbf{Llama-3-70B}}\\*
B & 87.10 & 59.00 & 88.50 & 96.17 & \textbf{96.33} & 95.50 & 97.27 & 93.00 & 98.17 & 98.33 & \textbf{98.83} & 98.00 \\
M & 90.97 & 72.67 & 93.67 & 95.67 & 94.67 & \textbf{98.17} & 98.03 & 95.50 & 98.83 & 97.83 & \textbf{99.00} & \textbf{99.00} \\
E & 89.97 & 65.00 & 91.83 & 97.33 & 97.50 & \textbf{98.17} & 96.63 & 94.00 & 97.67 & 97.00 & 96.50 & \textbf{98.00} \\
M$\times$2 & 85.60 & 52.50 & 85.17 & 95.83 & 95.67 & \textbf{98.83} & 97.57 & 92.83 & 98.50 & 98.83 & 98.67 & \textbf{99.00} \\
BM & 92.40 & 78.17 & 91.17 & \textbf{97.67} & 97.33 & \textbf{97.67} & 98.40 & 96.83 & 98.67 & 98.83 & 98.67 & \textbf{99.00} \\
BE & 97.20 & 94.67 & 98.00 & 96.83 & \textbf{98.67} & 97.83 & 99.13 & 98.50 & 99.33 & 99.50 & \textbf{99.67} & 98.67 \\
ME & 97.10 & 92.00 & 97.83 & 97.50 & 98.83 & \textbf{99.33} & 99.13 & 98.83 & \textbf{99.33} & 99.17 & 99.17 & 99.17 \\
M$\times$3 & 85.27 & 48.17 & 87.83 & 96.00 & 95.67 & \textbf{98.67} & 97.40 & 91.50 & 98.83 & 98.67 & 98.83 & \textbf{99.17} \\
BME & 96.77 & 92.33 & 95.83 & 98.00 & \textbf{98.83} & \textbf{98.83} & 99.07 & 98.67 & 99.17 & 99.17 & \textbf{99.50} & 98.83 \\
M$\times$5 & 86.10 & 50.17 & 90.00 & 95.83 & 95.67 & \textbf{98.83} & 96.70 & 88.17 & 98.67 & 98.67 & 98.67 & \textbf{99.33} \\
\hline
\multicolumn{13}{@{}l}{\textbf{Qwen3-8B}}\\*
B & 32.40 & 16.50 & 19.50 & \textbf{57.83} & 20.33 & 47.83 & 78.27 & 72.50 & 80.00 & \textbf{86.00} & 79.00 & 73.83 \\
M & 36.30 & 22.33 & 66.50 & 17.00 & 8.00 & \textbf{67.67} & 77.07 & 79.17 & \textbf{95.17} & 62.17 & 57.00 & 91.83 \\
E & 50.87 & 28.83 & 40.00 & 65.83 & 48.17 & \textbf{71.50} & 91.33 & 88.33 & 94.33 & 92.83 & 86.17 & \textbf{95.00} \\
M$\times$2 & 42.87 & 35.83 & 76.50 & 11.67 & 8.67 & \textbf{81.67} & 77.87 & 84.17 & \textbf{96.17} & 60.67 & 58.83 & 89.50 \\
BM & 51.70 & 27.33 & 30.33 & 73.17 & 48.67 & \textbf{79.00} & 87.00 & 83.00 & 86.17 & 86.33 & 86.50 & \textbf{93.00} \\
BE & 76.90 & 67.17 & 63.00 & 86.00 & \textbf{89.33} & 79.00 & 89.97 & 80.83 & 94.17 & 91.33 & \textbf{97.50} & 86.00 \\
ME & 72.60 & 68.33 & 90.00 & 63.67 & 47.83 & \textbf{93.17} & 88.90 & 95.50 & \textbf{98.67} & 77.33 & 76.00 & 97.00 \\
M$\times$3 & 40.37 & 34.33 & 70.67 & 10.33 & 5.67 & \textbf{80.83} & 76.00 & 81.33 & \textbf{96.00} & 61.67 & 53.33 & 87.67 \\
BME & 82.27 & 75.33 & 72.33 & \textbf{94.33} & 78.67 & 90.67 & 92.07 & 89.17 & 93.67 & 93.67 & 89.17 & \textbf{94.67} \\
M$\times$5 & 38.00 & 27.33 & 66.50 & 10.67 & 5.17 & \textbf{80.33} & 76.93 & 80.17 & \textbf{94.50} & 62.33 & 54.67 & 93.00 \\
\hline
\multicolumn{13}{@{}l}{\textbf{Qwen3-32B}}\\*
B & 80.13 & 73.00 & 68.83 & \textbf{93.67} & 86.50 & 78.67 & 93.97 & 95.33 & 96.00 & \textbf{98.33} & 91.50 & 88.67 \\
M & 84.73 & 79.00 & 87.00 & 87.83 & 80.00 & \textbf{89.83} & 92.13 & 92.67 & \textbf{98.33} & 89.67 & 85.00 & 95.00 \\
E & 94.97 & 92.83 & 90.83 & \textbf{97.83} & 96.17 & 97.17 & 97.63 & 93.83 & \textbf{99.33} & 98.83 & 98.67 & 97.50 \\
M$\times$2 & 83.27 & 84.83 & 85.83 & 81.33 & 71.33 & \textbf{93.00} & 91.47 & 93.67 & \textbf{99.50} & 87.33 & 80.00 & 96.83 \\
BM & 84.87 & 87.33 & 69.83 & \textbf{94.67} & 83.00 & 89.50 & 96.10 & 96.67 & 95.50 & \textbf{99.33} & 92.67 & 96.33 \\
BE & 96.53 & 97.00 & 94.67 & \textbf{98.67} & 97.83 & 94.50 & 97.60 & 97.83 & 98.67 & \textbf{99.33} & 95.83 & 96.33 \\
ME & 93.77 & 88.67 & 94.17 & \textbf{97.33} & 95.17 & 93.50 & 95.03 & 95.50 & \textbf{98.33} & 94.17 & 91.83 & 95.33 \\
M$\times$3 & 79.53 & 82.33 & 87.17 & 78.50 & 57.17 & \textbf{92.50} & 89.13 & 91.50 & \textbf{98.67} & 83.50 & 74.17 & 97.83 \\
BME & 91.03 & 92.83 & 84.67 & \textbf{97.33} & 89.00 & 91.33 & 96.13 & 96.83 & 96.33 & \textbf{98.83} & 91.00 & 97.67 \\
M$\times$5 & 77.87 & 80.17 & 86.17 & 77.67 & 54.00 & \textbf{91.33} & 89.33 & 93.17 & \textbf{98.50} & 83.67 & 74.00 & 97.33 \\
\hline
\caption{CK rates (\%) for every layout and model by sentence context. Each frame cell has 600 responses; All has 3,000. Within each context, \textbf{bold} marks the highest frame rate in the row.}\label{tab:paper-layout-full}\\
\end{longtable}
}
\twocolumn

\ifPDFTeX\end{CJK*}\fi
\end{document}